\documentclass[]{c4ggroup}
\usepackage{amsmath,amsfonts,bm}
\usepackage[T1]{fontenc}
\usepackage{booktabs}
\usepackage{multirow}
\usepackage{hyperref}
\usepackage{url}

\def\eqref#1{equation~\ref{#1}}

\def\1{\bm{1}}

\DeclareMathAlphabet{\mathsfit}{\encodingdefault}{\sfdefault}{m}{sl}
\SetMathAlphabet{\mathsfit}{bold}{\encodingdefault}{\sfdefault}{bx}{n}

\usepackage{xspace}
\newcommand{\methodname}{\textsc{NormGuard}\xspace}
\usepackage[table]{xcolor}
\usepackage{wrapfig}
\usepackage{caption}
\usepackage{inconsolata}
\usepackage[most]{tcolorbox}
\newtcblisting{promptbox}[1][]{
  listing only,
  enhanced jigsaw,
  breakable,
  listing options={breaklines=true,breakautoindent=false,breakindent=0pt,basicstyle=\ttfamily\scriptsize,
    literate={–}{{\textendash}}1 {"}{{\textquotedbl}}1},
  colback=tablegray,
  colframe=goodcolor,
  boxrule=0.4pt,
  arc=2mm,
  leftrule=0.4pt,
  rightrule=0.4pt,
  top=4pt,
  bottom=4pt,
  left=6pt,
  right=6pt,
  #1
}
\usepackage{fvextra}
\usepackage{tikz}
\usepackage{enumitem}
\newtheorem{definition}{Definition}

\usepackage{pifont}
\newcommand{\cmark}{\ding{51}}
\newcommand{\xmark}{\ding{55}}
\usetikzlibrary{positioning}
\definecolor{radialyellow}{RGB}{255,255,180}
\definecolor{angularblue}{RGB}{200,220,255}
\definecolor{tablegray}{RGB}{239,248,249}
\definecolor{goodcolor}{HTML}{3359B6}
\definecolor{badcolor}{HTML}{3FBCC6}
\usepackage{soul}
\sethlcolor{radialyellow}

\title{\methodname: Reward-Preserving Norm Constraints in Flow-Matching Reinforcement Learning}

\author[1,2,3]{Tianlin Pan\textsuperscript{*}}
\author[1,2]{Lianyu Pang\textsuperscript{*}}
\author[2]{Cheng Da}
\author[2]{Huan Yang}
\author[2]{Changqian Yu}
\author[2]{Kun Gai}
\author[1]{Wenhan Luo\textsuperscript{\dag}}

\affiliation[1]{The Hong Kong University of Science and Technology}
\affiliation[2]{KlingAI Research}
\affiliation[3]{University of Chinese Academy of Sciences}

\begin{document}

\abstract{Reinforcement learning (RL) post-training improves the reward alignment of flow-based generators, but often degrades perceptual quality in ways that are not captured by the reward proxy.
We identify a simple structural signature of this drift: across three post-training methods (NFT, AWM, DPO), RL fine-tuning inflates the per-step velocity norm $\|v_\theta\|$ by $5\%$ to $15\%$ relative to the reference. This inflation is perceptually consequential: because the generated sample is the displacement accumulated by the velocity field, a $\delta\%$ norm inflation raises the second moment of the sample by roughly $2\delta\%$, which surfaces in the decoded image as deviations in luminance and chrominance, i.e., boosted contrast, over-sharpened edges, and unnatural white balance. The obvious remedy, rescaling $v_\theta$ back to $\|v_{\text{ref}}\|$ at inference time as done for classifier-free guidance, does not transfer to RL: it neither improves reward nor fixes the quality degradation, because the inflation is co-adapted into the model weights. We further show that velocity magnitude rescaling carries no coherent reward signal at the batch level, indicating that suppressing norm inflation is unlikely to remove a consistently reward-carrying component. Both findings point to \textbf{\methodname}, a simple hinge penalty that activates only when $\|v_\theta\|$ exceeds $\|v_{\text{ref}}\|$ and composes additively with any velocity-local base loss. Across two base models, three post-training methods, and two reward proxies, \textbf{\methodname} consistently improves MLLM-judged image quality, usually improves forensic realism, and largely preserves reward, with gains that amplify under few-step inference and are not explained by early stopping.}

\maketitle

\begin{figure*}[h]
\centering
\includegraphics[width=\linewidth]{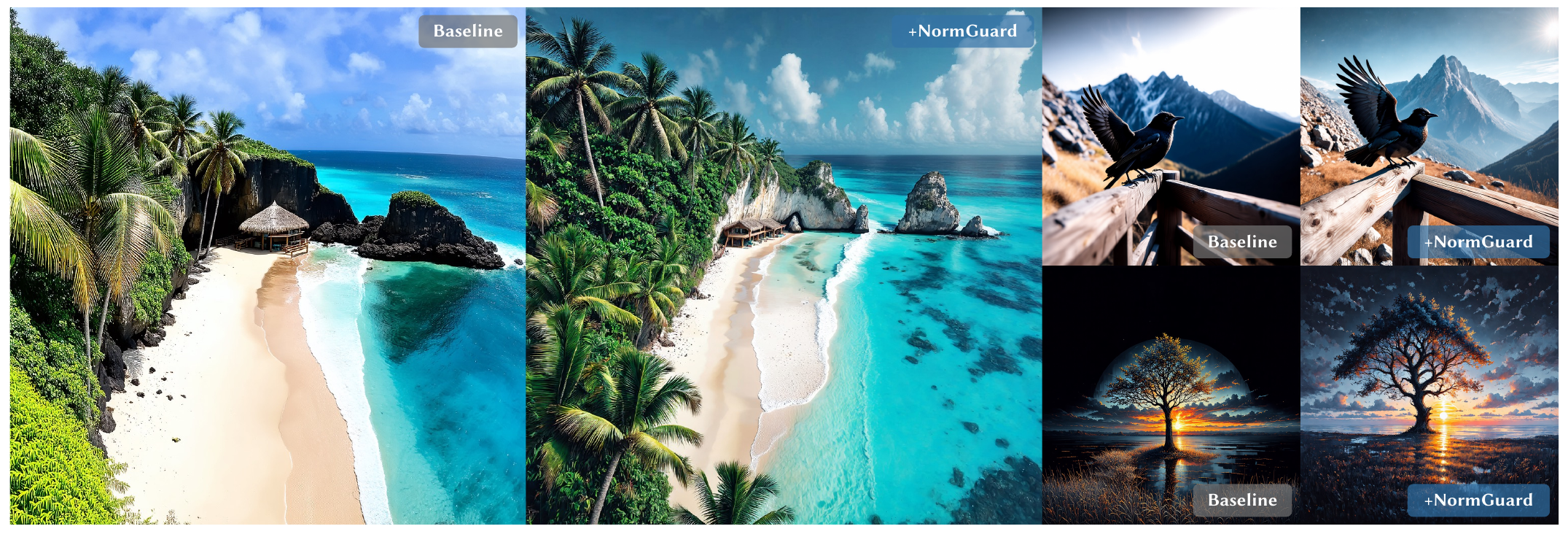}
\vspace{-20pt}
\caption{We propose \textbf{\methodname}, a simple norm-budget regularizer that suppresses
reward-irrelevant norm inflation during RL training of flow-matching models.
\methodname reduces the over-sharpening, color
oversaturation, and unnatural lighting seen in the baseline,
producing more photo-realistic results. Prompts see Appendix~\ref{sec:appendix-prompts-teaser}.}
\label{fig:teaser}
\end{figure*}

\section{Introduction} \label{sec:introduction}

Reinforcement learning (RL) post-training has become a standard tool for aligning flow-based generative models~\cite{ddpm,ddim,ldm,flow-matching} with human preferences~\cite{PickScore,HPSv2}. Reward gains, however, are consistently accompanied by \emph{reward over-optimization}~\cite{ScalingRewardOveroptimization,realgen,GARDO}: perceptual quality degrades in ways that are not captured by the reward proxy, including over-sharpening, color shift, unnatural lighting, and loss of fine texture. The standard mitigations, such as early stopping and KL regularization, treat the post-training drift as a single aggregate quantity. Let $v_\theta$ denote the fine-tuned velocity and $v_{\text{ref}}$ the pre-trained reference. KL regularization typically takes the form of an MSE penalty $\|v_\theta - v_{\text{ref}}\|^2$. These methods constrain how much the velocity deviates in total but do not distinguish which component of that deviation is associated with the artifact, leaving no basis for a targeted fix.

\begin{figure}[t]
\centering
\includegraphics[width=\linewidth]{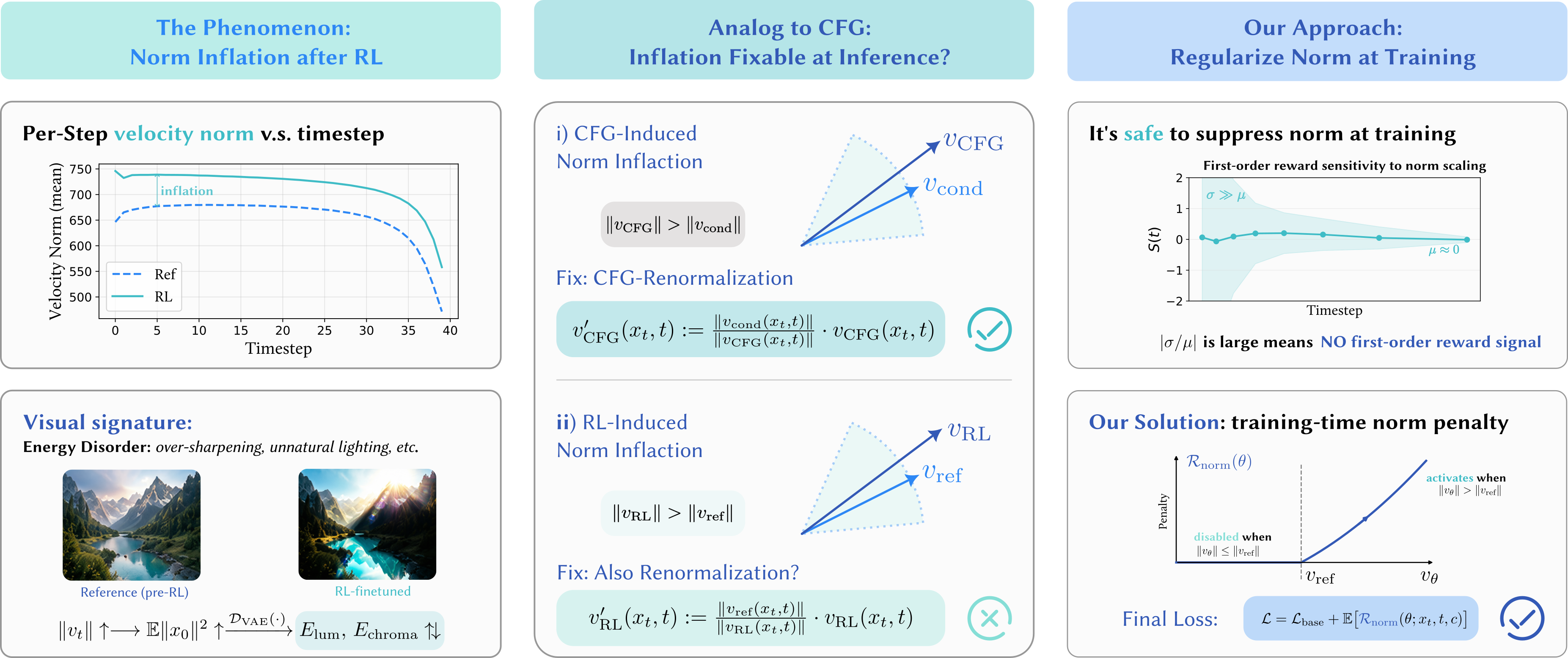}
\vspace{-10pt}
\definecolor{radialyellow}{RGB}{203,203,2}
\definecolor{angularblue}{RGB}{49,154,255}
\caption{\textbf{Motivation}: RL post-training inflates the per-step velocity norm and produces visual artifacts (\textbf{left}). Unlike CFG, inference-time renormalization fails for RL because the inflation is co-adapted into the model weights (\textbf{middle}). An
adjoint sensitivity analysis confirms that suppressing norm inflation carries no coherent first-order reward signal, motivating our training-time hinge penalty on excess velocity norm (\textbf{right}).}

\label{fig:idea}
\end{figure}

Figure~\ref{fig:idea} outlines the full motivation. Without imposing any structural assumption on the drift, a simple starting point is to inspect the per-step velocity norm $\|v_\theta(x_t, t)\|$ and its ratio to the reference norm $\|v_{\text{ref}}(x_t, t)\|$. Take SD3.5-Medium with PickScore as an example: across NFT, AWM, and DPO, RL post-training consistently inflates this norm by $5\%$ to $15\%$ relative to the reference, uniformly along the denoising trajectory (Figure~\ref{fig:norm_inflation}). This is not merely an internal statistic. Because the generated latent is the displacement accumulated by the velocity field, scaling that field by $1+\delta$ along the trajectory raises the second moment of the generated sample by roughly $2\delta$, and the surplus energy is empirically accompanied by deviations in decoded-image luminance and chrominance statistics, i.e., the boosted contrast, over-sharpened edges, and unnatural white balance observed in Figure~\ref{fig:teaser}.

A similar form of norm inflation has been well documented in inference-time classifier-free guidance (CFG)~\cite{cfg}. For guidance scale $\omega>1$, the CFG-modified velocity has a substantially larger magnitude than the conditional velocity ($\|v_{\text{CFG}}\| > \|v_{\text{cond}}\|$), and this norm growth has been shown to drive sampling trajectories to overshoot beyond the learned data distribution~\cite{cfg-renorm}. As shown in Figure~\ref{fig:idea} (middle), \cite{cfg-renorm} address the CFG case by rescaling the deployed velocity back to a reference norm at inference time while preserving its direction, and this simple fix successfully resolves the CFG artifacts. 

However, this CFG-renormalization technique does not apply to the norm inflation induced by RL post-training. We scale the RL-fine-tuned velocity back to match $\|v_{\text{ref}}\|$ at every step. Reward is essentially unchanged after renormalization, yet the resulting images exhibit over-sharpening and unnatural lighting (Table~\ref{tab:velo_clip} and Figure~\ref{fig:renorm-artifacts}). The contrast with CFG is diagnostic: CFG inflation is an explicit inference-time combination of two velocity heads, so rescaling removes it cleanly; RL inflation, by contrast, is trained into the model weights, and rescaling it back at inference distorts the co-adapted dynamics.

Then, we ask whether suppressing the inflation during training would disturb reward gains. The first-order reward change under a multiplicative velocity scaling $v_\theta \mapsto (1+\varepsilon)v_\theta$ is governed by a per-timestep norm-scaling sensitivity $S(t) = v_\theta(x_t, t)^\top a(t)$, where $a(t)$ is the reward adjoint state. On $6{,}400$ samples, $S(t)$ exhibits substantial per-sample magnitude and sign heterogeneity across prompts, while its batch mean remains close to zero; the noise-to-signal ratio $\sigma/|\mu|$ ranges from $3\times$ to $100\times$. Thus, at first order and at the batch level, velocity magnitude rescaling does not reveal a coherent reward signal, suggesting that suppressing norm inflation is unlikely to remove reward gains at training.

The two findings point in the same direction. \textbf{1)} Inference-time renormalization fails, so the intervention must happen at training time. \textbf{2)} Velocity magnitude rescaling does not reveal a coherent first-order reward signal at the batch level, suggesting that training-time norm suppression is unlikely to systematically reduce reward gains. Based on these observations, we propose \textbf{\methodname}, a training-time penalty on the velocity norm that activates only when $\|v_\theta\|$ exceeds $\|v_{\text{ref}}\|$, as shown in Figure~\ref{fig:idea} (right). The objective sits alongside the existing velocity-local post-training loss and only adds a single scalar hyperparameter.

We validate this regularizer across two base flow models, three post-training methods (NFT, AWM, DPO), and two reward models. Using both MLLM-based image quality assessment and forensic realism detection, we find that the regularizer consistently improves visual quality and, in most settings, improves realism while preserving the majority of the reward gain. The image-space measurements confirm the energy chain above: RL post-training inflates the luminance energy by $24\%$ to $46\%$ and shifts the chroma energy away from the pretrained reference, and our regularizer reduces both deviations. Notably, the gains are particularly pronounced at few-step inference, and the improvement cannot be explained by early stopping and is complementary to KL regularization.

Our contributions are threefold. \textbf{1)} We identify a consistent velocity-norm inflation effect in RL post-training for flow-based generative models, and give an energy account of why it is perceptually visible, tying it to measurable excess luminance and chroma energy in the decoded image. \textbf{2)} We show that a CFG-style inference-time renormalization does not transfer cleanly to RL-trained models, while batch-level first-order sensitivity does not reveal a stable reward effect aligned with uniform norm scaling. \textbf{3)} We introduce \textbf{\methodname}, a simple training-time regularizer that suppresses excess velocity-norm growth and improves perceptual quality across a range of models, post-training methods, and reward functions.

\section{Related Work}
\subsection{RL Post-Training of Generative Models}
Text-to-image generation has seen significant improvements in recent years~\cite{ddpm,ldm,SD3,flux-2}. Inspired by the success of RLHF in language models~\cite{ppo,dpo,grpo}, a growing body of work has adapted these techniques to flow-based generative models, including Diffusion-DPO~\cite{DiffusionDPO}, DDPO~\cite{DDPO}, Flow-GRPO~\cite{Flow-grpo}, Dance-GRPO~\cite{DanceGRPO}, Diffusion-NFT~\cite{NFT}, and AWM~\cite{AWM}. All of these methods rely on a reward model such as PickScore~\cite{PickScore}, GenEval~\cite{GenEval}, or HPS~\cite{HPSv2,HPSv3} to provide the training signal.
\subsection{Reward Over-optimization}
A well-documented pathology of RL fine-tuning is reward over-optimization~\cite{ScalingRewardOveroptimization,ConfrontingRewardOveroptimization,DefiningRewardGaming}, where models exploit imperfections
in the reward proxy at the expense of true quality. In text-to-image generation, optimizing rewards such as PickScore or HPS can produce artifacts including over-sharpening, color bias, and unnatural
lighting~\cite{realdpo,realgen}, so proxy reward may improve even as perceptual quality deteriorates.

Several recent works address this issue from different angles. GRPO-Guard~\cite{GRPO-Guard} stabilizes optimization with ratio normalization and gradient reweighting. RSA-FT~\cite{rsa-ft} attributes
reward hacking to sharp reward landscapes and mitigates it by flattening the landscape in image and parameter space. RealGen~\cite{realgen} replaces human-preference rewards with detector-based rewards
better aligned with photorealism. RewardDance~\cite{RewardDance} reduces reward hacking by scaling up the reward model, Pref-GRPO~\cite{Pref-GRPO} replaces pointwise score maximization with pairwise
preference fitting to avoid illusory advantages, and GARDO~\cite{GARDO} introduces gated adaptive regularization and diversity-aware optimization to balance anti-hacking and exploration.
\section{Norm Inflation: Phenomenon and Diagnostics} \label{sec:motivation}

This section sharpens the empirical motivation for our method. We first document the norm inflation induced by RL post-training, explain through an energy argument why it is perceptually visible, and relate it to a similar phenomenon in classifier-free guidance (Section~\ref{sec:phenomenon}). We then examine whether the corresponding inference-time renormalization transfers to RL-trained models, and whether suppressing the inflated norm exhibits a stable first-order reward effect at the batch level (Section~\ref{sec:diagnostics}).

\subsection{The Phenomenon: RL-Induced Velocity Norm Inflation} \label{sec:phenomenon}

\begin{figure}[t]
\begin{minipage}[t]{0.56\textwidth}
\centering
\includegraphics[width=\linewidth]{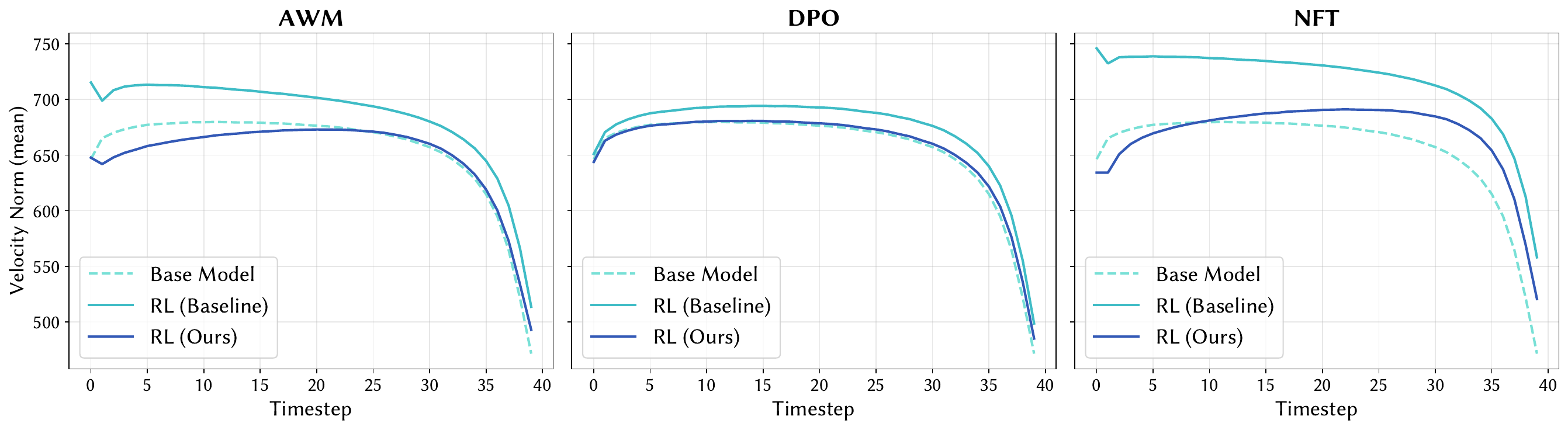}
\captionof{figure}{\textbf{RL-induced norm inflation.} Mean velocity norm across denoising timesteps for AWM, DPO, and NFT on SD3.5-Medium. RL post-training uniformly inflates the norm above the pretrained reference (dashed). \textbf{\methodname} regularizer suppresses the excess without sacrificing reward.}
\label{fig:norm_inflation}
\end{minipage}
\hfill
\begin{minipage}[t]{0.40\textwidth}
\centering
\includegraphics[width=\linewidth]{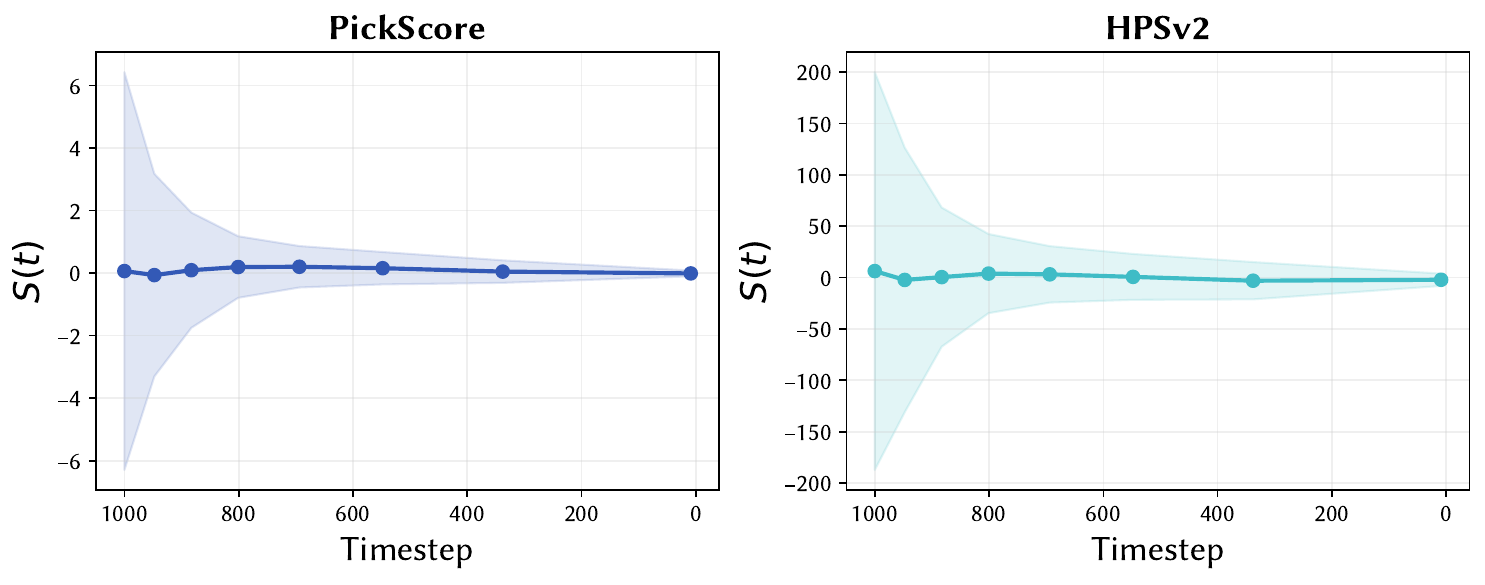}
\captionof{figure}{\textbf{Norm-Scaling sensitivity} $S(t)$ across denoising steps (SD3.5-M + 6.4K samples). The noise-to-signal ratio $\sigma/|\mu|$ ranges from $3\times$ to $100\times$, indicating noisy signal along the norm inflation direction.}
\label{fig:radial-sensitivity}
\end{minipage}
\end{figure}

We measure the per-step velocity norm $\|v_\theta(x_t, t)\|$ before and after RL post-training on SD3.5-Medium + PickScore, across three RL methods (NFT~\cite{NFT}, AWM~\cite{AWM} and DPO~\cite{DiffusionDPO}). RL fine-tuning consistently shifts the velocity norm distribution upward by $5\%$ to $15\%$, uniformly along the denoising trajectory and across all configurations (Figure~\ref{fig:norm_inflation}). The corresponding samples share a characteristic look: over-sharpened edges, boosted contrast, and unnatural white balance.

\paragraph{From velocity norm to image energy.}
These two observations are linked by an energy budget. Let $\epsilon \sim \mathcal{N}(0, I)$ be the initial noise and $x_0 = \epsilon - \int_0^1 v_\theta(x_t, t)\,\mathrm{d}t$ the generated latent, so that $D := \epsilon - x_0$ is the total displacement produced by the velocity field. Consider the idealized form of the inflation measured above: the field is scaled by a uniform factor $1+\delta$ along the trajectory with its direction unchanged. Then $D \mapsto (1+\delta) D$ and the endpoint moves to $x_0' = \epsilon - (1+\delta) D = (1+\delta)\,x_0 - \delta\,\epsilon$. Since the initial noise is essentially uncorrelated with the generated latent ($\mathbb{E}\langle x_0, \epsilon \rangle \approx 0$),
\begin{equation}
  \mathbb{E}\|x_0'\|_2^2 = (1+\delta)^2\,\mathbb{E}\|x_0\|_2^2 - 2\delta(1+\delta)\,\mathbb{E}\langle x_0, \epsilon\rangle + \delta^2\,\mathbb{E}\|\epsilon\|_2^2 \approx (1 + 2\delta)\,\mathbb{E}\|x_0\|_2^2 + O(\delta^2),
\end{equation}
i.e., a $\delta$-level norm inflation raises the second moment of the generated latent by roughly $2\delta$. This surplus energy is then passed on to the decoded image as excess variation around its mean.

\paragraph{Image-space energy statistics.}
We convert decoded images to the standard $\mathrm{YC_bC_r}$ space and record the per-image spatial second central moments of the luminance and chroma channels, $E_{\text{lum}} := \operatorname{Var}[Y]$ and $E_{\text{chroma}} := \operatorname{Var}[C_b] + \operatorname{Var}[C_r]$ (details in Appendix~\ref{sec:appendix-energy-stats}). By Parseval's identity, $E_{\text{lum}}$ is exactly the total luminance energy carried by all non-DC spatial frequencies, i.e., how much structure the image places above its mean brightness, and $E_{\text{chroma}}$ is its color counterpart. As shown in Figure~\ref{fig:quality-energy}, norm inflation drives both statistics away from the pretrained reference. In the visual samples of Figure~\ref{fig:teaser}, excess $E_{\text{lum}}$ appears as boosted contrast and over-sharpened edges (crushed shadows, blown highlights), while a shifted $E_{\text{chroma}}$ appears as over-saturation or unnatural white balance.

\paragraph{CFG-induced velocity inflation.}
The same pairing of inflated velocity norm and these artifacts is documented for inference-time classifier-free guidance (CFG)~\cite{cfg}. STIV~\cite{cfg-renorm} observe that for guidance scale $\omega>1$ the guided velocity $\hat{v}_{\text{CFG}} = v_{\text{uncond}} + \omega(v_{\text{cond}} - v_{\text{uncond}})$ satisfies $\|\hat{v}_{\text{CFG}}\| > \|v_{\text{cond}}\|$, which drives the sampling trajectory to overshoot beyond the learned data distribution. Their fix, \emph{CFG-Renormalization}, removes the artifacts at inference time by rescaling $\hat{v}_{\text{CFG}}$ back to $\|v_{\text{cond}}\|$ while preserving its direction. The next subsection asks whether the same intervention transfers to an RL-trained model.

\subsection{Can Inference-Time Renormalization Transfer to RL?} \label{sec:diagnostics}

\begin{figure}[ht]
\begin{minipage}[t]{0.52\textwidth}
\centering
\captionof{table}{\textbf{Inference-time renormalization does not improve reward.} $\Delta R$ denotes the change in reward w/ and w/o velocity renormalization.}
\label{tab:velo_clip}
\renewcommand{\arraystretch}{1.2}
\setlength{\tabcolsep}{4pt}
\scriptsize
\begin{tabular}{llcccc}
\toprule
Metric & Config. & Bas. & RL & +Renorm & $\Delta R$ \\
\midrule
\multirow{4}{*}{\shortstack[l]{Pick-\\Score}} & SD3.5 (NFT) & 0.768 & 0.869 & 0.869 & \cellcolor{tablegray}{\textcolor{goodcolor}{$+$\textbf{0.000}}} \\
                           & SD3.5 (AWM) & 0.768 & 0.869 & 0.865 & \cellcolor{tablegray}{\textcolor{badcolor}{$-$\textbf{0.004}}} \\
                           & SD3.5 (DPO) & 0.768 & 0.849 & 0.850 & \cellcolor{tablegray}{\textcolor{goodcolor}{$+$\textbf{0.001}}} \\
                           & FLUX (NFT)  & 0.802 & 0.905 & 0.904 & \cellcolor{tablegray}{\textcolor{badcolor}{$-$\textbf{0.001}}} \\
\midrule
\multirow{2}{*}{HPSv2} & SD3.5 (NFT) & 0.289 & 0.308 & 0.300 & \cellcolor{tablegray}{\textcolor{badcolor}{$-$\textbf{0.008}}} \\
                       & FLUX (NFT)  & 0.271 & 0.315 & 0.313 & \cellcolor{tablegray}{\textcolor{badcolor}{$-$\textbf{0.002}}} \\
\bottomrule
\end{tabular}

\end{minipage}
\hfill
\begin{minipage}[t]{0.44\textwidth}
\centering
\vspace{1pt}
\includegraphics[width=0.9\linewidth]{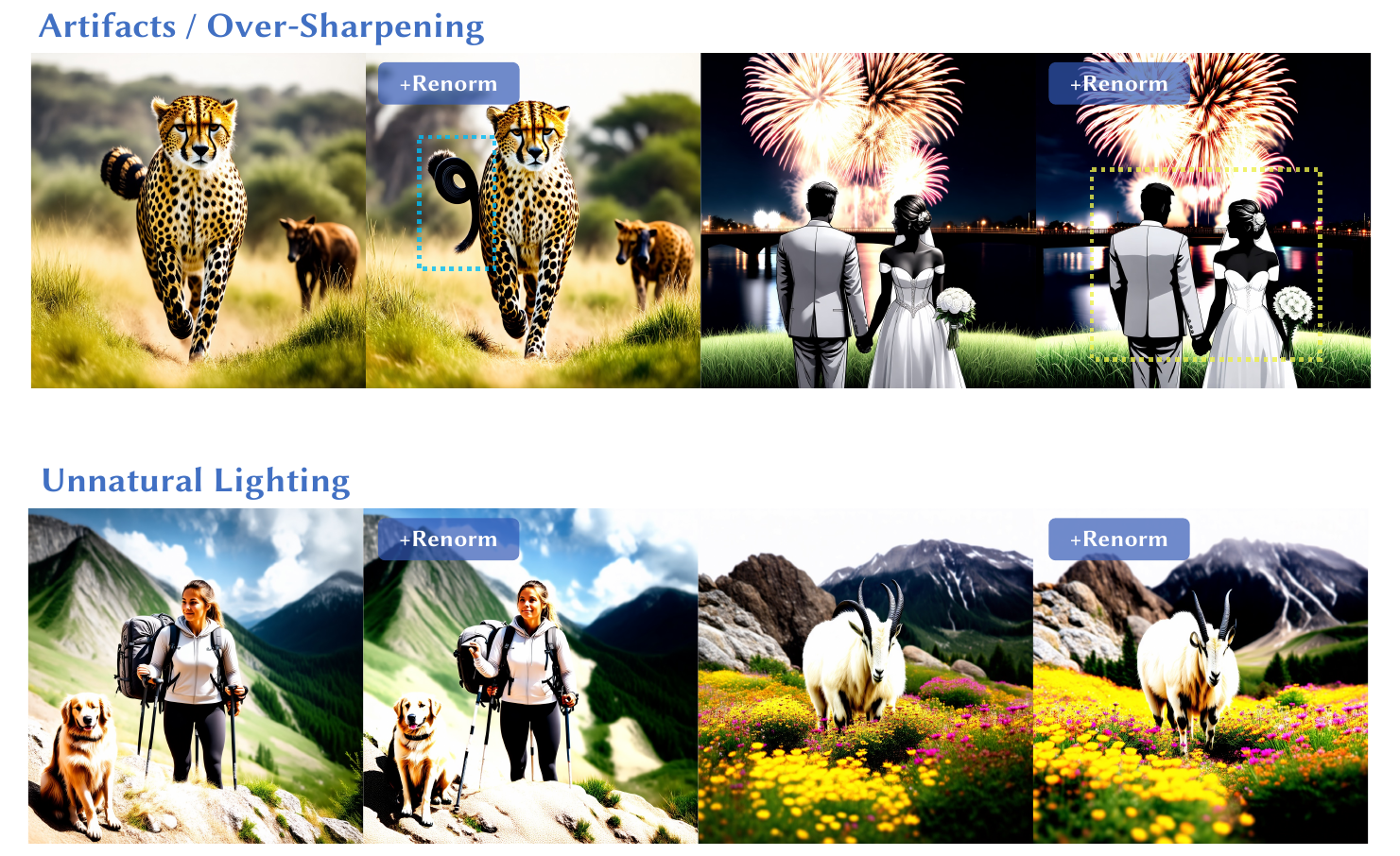}
\captionof{figure}{\textbf{Inference-time renormalization introduces additional artifacts.}}
\label{fig:renorm-artifacts}
\end{minipage}
\end{figure}

We present two measurements. Observation~1 shows that the inference-time correction used in CFG does not transfer cleanly to the RL setting. Observation~2 examines whether suppressing velocity magnitude exhibits a stable first-order reward effect at the batch level. 

\paragraph{Observation 1: Inference-time renormalization does not transfer cleanly to RL-trained models.}
We test whether the CFG-renorm-style inference fix transfers. Given a reward-fine-tuned $v_\theta$, we apply the per-step inference-time norm scaling
\begin{equation}
  v_\theta'(x_t, t) := {\|v_{\text{ref}}(x_t, t)\|} / {\|v_\theta(x_t, t)\|} \cdot v_\theta(x_t, t),
\end{equation}
which preserves the direction of $v_\theta$ and rescales its magnitude to the reference norm. Table~\ref{tab:velo_clip} and Figure~\ref{fig:renorm-artifacts} report the result. Reward does not improve after renormalization, and the sampled images exhibit over-sharpening and unnatural lighting. This differs from the CFG case, where norm inflation is introduced explicitly at inference time and can be removed by renormalizing the deployed velocity. In the RL setting, the same intervention appears insufficient once the inflated norm has been absorbed by the fine-tuned model, suggesting that a more effective intervention should be applied during training rather than only at inference time.

\paragraph{Observation 2: Velocity rescaling does not exhibit a stable reward effect at the batch level.}
Independently of Observation~1, we ask whether suppressing the inflated norm would directly remove useful reward signal. Keeping the convention of Section~\ref{sec:phenomenon}, where the sampler integrates from the noise end $t=1$ down to the generated sample $x_0$, the first-order reward change under a velocity-space perturbation $v_\theta \mapsto v_\theta + \varepsilon\,\eta$ is
\begin{equation}
  \frac{\mathrm{d}R}{\mathrm{d}\varepsilon}\Big|_{\varepsilon=0} = -\int_0^1 a(t)^\top \eta(x_t, t)\,\mathrm{d}t,
\end{equation}
where the adjoint state $a(t) := \partial R / \partial x_t$ satisfies $\dot{a}(t) = -[\nabla_{x_t} v_\theta]^{\top} a(t)$ with boundary condition $a(0) = \nabla_{x_0} R(x_0)$ at the generated sample. Setting $\eta = v_\theta$ (i.e., a multiplicative magnitude scaling $v_\theta \mapsto (1+\varepsilon) v_\theta$) yields the \emph{norm-scaling sensitivity} $S(t) := v_\theta(x_t, t)^{\top} a(t)$, which measures the marginal first-order reward effect of rescaling velocity magnitude at time $t$.

We compute $S(t)$ on $6{,}400$ samples from SD3.5-Medium with PickScore and HPSv2. Across all timesteps, $S$ exhibits substantial per-sample magnitude and sign-heterogeneous behavior across prompts, while its batch mean remains close to zero; the noise-to-signal ratio $\sigma/|\mu|$ ranges from $3\times$ to $100\times$ (Figure~\ref{fig:radial-sensitivity}).
These measurements do not reveal a stable batch-level first-order reward signal aligned with uniform norm scaling. This suggests that suppressing the inflated norm is unlikely to systematically remove a reward-carrying direction.

Taken together, these observations point to a training-time solution. Direct inference-time renormalization is insufficient in the RL setting, while the sensitivity analysis does not reveal a stable batch-level first-order reward effect tied to uniform norm scaling. This motivates the regularizer introduced next, which suppresses excess velocity-norm growth during post-training.

\section{\methodname} \label{sec:method}

We now present \methodname. We first identify a local output-space structure
shared by the post-training losses considered in this paper
(Section~\ref{sec:method-velocity-local}), and then introduce a norm-budget
regularizer that operates in the same space
(Section~\ref{sec:method-regularizer}).

\subsection{Velocity-Local Post-Training Losses}
\label{sec:method-velocity-local}

We focus on post-training objectives whose gradients act through local velocity
residuals in the same output space as flow matching. For a noisy state $x_t$ at
timestep $t$ under condition $c$, let $v_\theta(x_t, t, c)$ denote the model
velocity and let $\tilde{v}(x_t, t, c)$ denote a local target velocity. The following definition formalizes this shared structure.

\begin{definition}[Velocity-local post-training loss]
\label{def:velocity-local}
A post-training objective $\mathcal{L}_{\text{post}}$ is \emph{velocity-local} if
its parameter gradient admits the form
\begin{equation}
  \nabla_\theta \mathcal{L}_{\text{post}}
  = \mathbb{E}\!\left[
      w(x_t, t, c)\, \nabla_\theta \|v_\theta(x_t, t, c) - \tilde{v}(x_t, t, c)\|_2^2
    \right],
\end{equation}
where $w(x_t, t, c)$ is a scalar weight that may depend on the sampled state $x_t$, timestep $t$, condition $c$, or reward label. \textnormal{This definition isolates whether the update is driven by per-timestep residuals in the velocity output space.}
\end{definition}

The three post-training methods used in our experiments fit this template.
\textbf{NFT} uses reward-weighted flow-matching residuals to a target velocity.
\textbf{AWM} reduces, under the FM-ELBO surrogate, to an advantage-weighted
flow-matching residual with a KL term instantiated as
$\|v_\theta - v_{\text{ref}}\|_2^2$. \textbf{DPO} injects preference signal
through sigmoid-weighted local deviations from the reference velocity on
winner--loser pairs. By contrast, \textbf{Flow-GRPO} is a non-example: its
gradient passes through a trajectory-level likelihood ratio over reverse
transitions, rather than a reweighting of single-step velocity residuals. Full
derivations are deferred to Appendix~\ref{sec:appendix-velocity-local}.

\subsection{\methodname: Norm Regularizer}
\label{sec:method-regularizer}

Our goal is to suppress excess velocity-norm growth during post-training while
interfering as little as possible with directional updates. We therefore add a
one-sided penalty that activates only when the current velocity norm exceeds the
reference-model norm.

\paragraph{Regularizer.}
For each sampled $(x_t, t, c)$, we define
\begin{equation}
  \mathcal{R}_{\text{norm}}(\theta; x_t, t, c)
  = \lambda \cdot \max\!\left\{0,\;
    \frac{\|v_\theta(x_t, t, c)\|_2^2 - \|v_{\text{ref}}(x_t, t, c)\|_2^2}{\|v_{\text{ref}}(x_t, t, c)\|_2^2}\right\},
\end{equation}
and optimize the batch-averaged objective
\begin{equation}
  \mathcal{L} = \mathcal{L}_{\text{base}} + \mathbb{E}\big[\mathcal{R}_{\text{norm}}(\theta; x_t, t, c)\big],
\end{equation}
where $\lambda > 0$ controls the regularization strength. The hinge leaves
updates unconstrained as long as $\|v_\theta\|_2^2 \leq
\|v_{\text{ref}}\|_2^2$, and penalizes only the excess norm beyond this
reference budget.

\paragraph{Compatibility with velocity-local losses.}
The penalty acts in the same velocity output space as the base loss. Whenever
the hinge is active, its gradient is proportional to $J_\theta^{\top}v_\theta$,
so it modifies the update through the local velocity at the sampled state rather
than through a separate parameter-space constraint. As a result,
$\mathcal{R}_{\text{norm}}$ composes naturally with any velocity-local base loss,
including the NFT, AWM, and DPO objectives.

\section{Experiments}
\label{sec:experiments}

\begin{figure}[ht]
    \centering
    \begin{minipage}[t]{0.38\textwidth}
        \centering
        \vspace{0cm}
        \includegraphics[width=\textwidth]{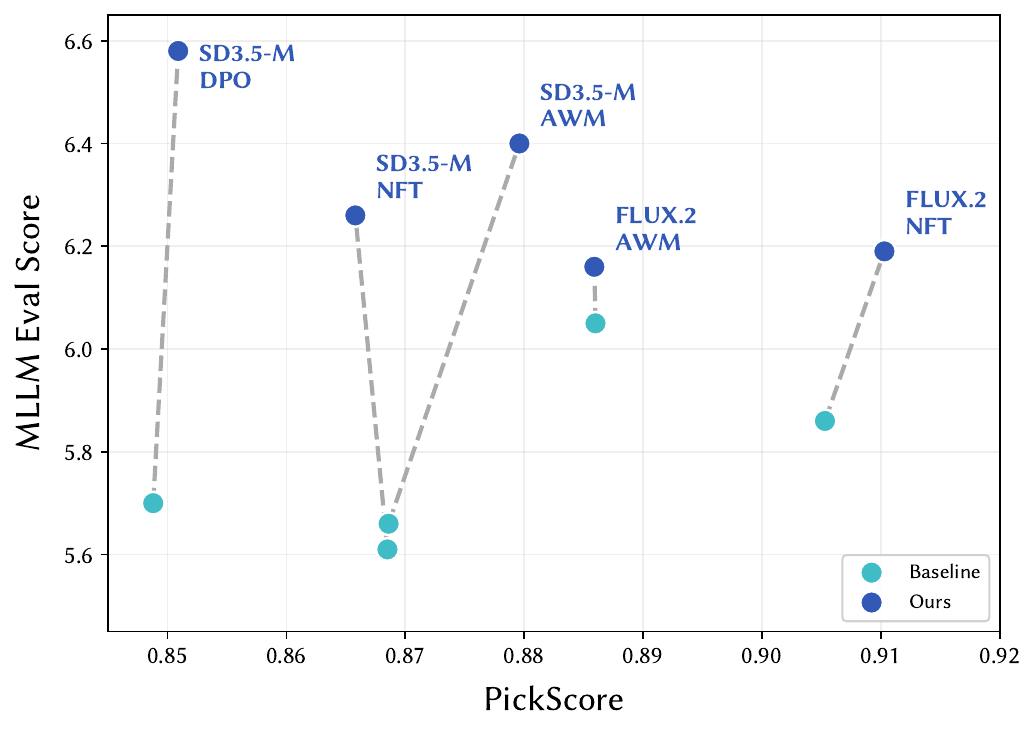}
        \vspace{-15pt}
        \captionof{figure}{\textbf{Reward vs.\ MLLM quality score.} \textbf{\methodname} improves quality with minimal reward change.}
        \label{fig:reward_vs_quality}
    \end{minipage}%
    \hfill
    \begin{minipage}[t]{0.60\textwidth}
        \centering
        \vspace{0cm}
        \includegraphics[width=\textwidth]{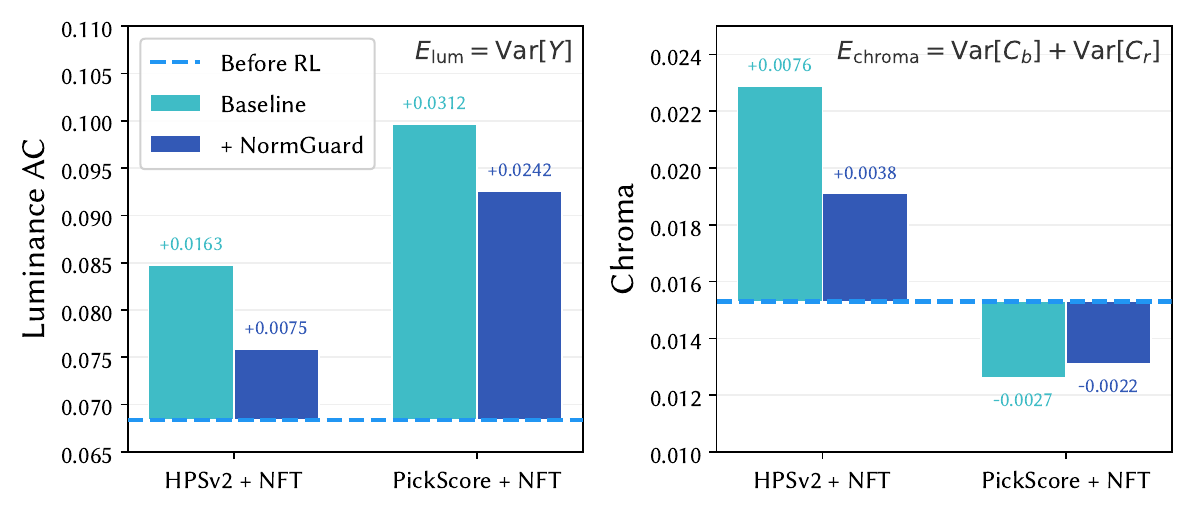}
        \vspace{-15pt}
        \captionof{figure}{\textbf{Image-Space Statistics.} Norm inflation deviates  energy-related statistics from the reference, while \textbf{\methodname} suppresses the deviation, leading to more natural luminance and color variation. Details see Appendix~\ref{sec:appendix-energy-stats}.}
        \label{fig:quality-energy}
    \end{minipage}
\end{figure}

\subsection{Evaluation Benchmarks}

\paragraph{Experimental Setup.}
We evaluate our \textbf{\methodname} regularizer across two base flow models (SD3.5-Medium~\cite{SD3} and FLUX.2-klein-base-4B~\cite{flux-2}), three RL post-training methods (NFT~\cite{NFT}, AWM~\cite{AWM}, DPO~\cite{DiffusionDPO}), and two reward models (PickScore~\cite{PickScore}, HPSv2~\cite{HPSv2}). We conduct all experiments via Flow-Factory~\cite{Flow-Factory} on 8$\times$ GPUs. Training prompts are drawn from the PickScore dataset, and test prompts are drawn from HPDv3~\cite{HPSv3}. Training details are provided in Appendix~\ref{sec:appendix-training-details}.

\paragraph{MLLM Image Quality Assessment.}
Following recent work on high-quality RL~\citep{realdpo, realgen}, we use multimodal LLMs as judges to assess image \emph{quality}. We employ both Qwen3.5-35B-A3B~\cite{qwen3.5} and GPT-4.1~\cite{gpt4.1} to score generated images on six axes: physical plausibility, texture and material fidelity, edge and boundary coherence, color and tone consistency, semantic coherence, and artifact detection (detailed prompts see Appendix~\ref{sec:appendix-mllm-prompts}). For each prompt, both methods generate an image and the judge provides a pairwise preference (Win / Tie / Loss) as well as an absolute score ($1\sim10$ scale). For each configuration, the rates are evaluated over $3$ seeds with $300$ samples, where the average rates are reported.
\paragraph{Forensic Realism Detection.}
Following RealGen~\cite{realgen}, we additionally evaluate with Forensic-Chat~\cite{ForensicChat}, an AIGC detector that scores how closely a generated image resembles a real photograph versus a synthetic output. This provides a complementary measure of \emph{realism} independent of aesthetic preference, capturing artifacts such as unnatural lighting and over-sharpening that reward models may overlook.

\subsection{Main Results}

\begin{figure}[ht]
\centering
\includegraphics[width=\linewidth]{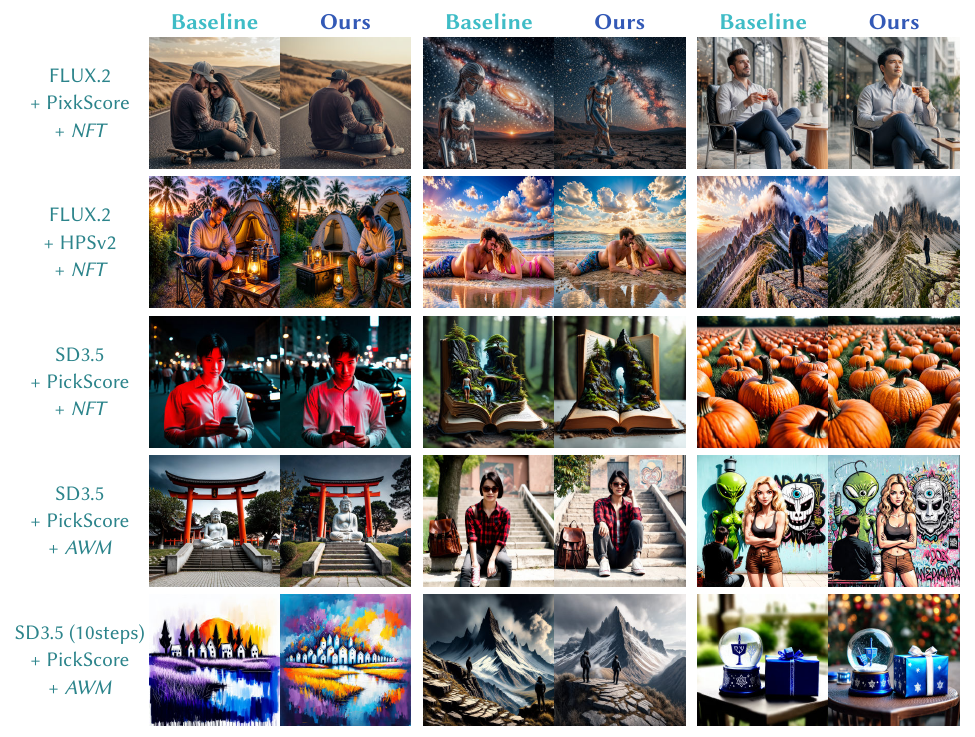}
\vspace{-10pt}
\caption{\textbf{Qualitative results.} We compare samples with and without \textbf{\methodname} regularizer. Prompts see Appendix~\ref{sec:appendix-prompts-main-results}.}
\label{fig:qual_results}
\end{figure}

\paragraph{Qualitative Results.}
Figure~\ref{fig:qual_results} shows representative samples from our method and the unregularized baseline across configurations. The baseline images exhibit a range of quality issues, including texture degradation, edge blurring, and artifact introduction. In contrast, our regularized models produce natural edges, more coherent textures, and fewer artifacts, suggesting that suppressing radial inflation is associated with tangible quality improvements in the generated images.

\paragraph{Image Quality and Realism.}
\begin{table}[t]
\scriptsize
\renewcommand{\arraystretch}{1.2}
\caption{\textbf{Image Quality \& Realism}. \textbf{\methodname} improves MLLM-judged quality (Qwen3.5 + GPT-4.1) and forensic realism (RealScore) while largely retaining reward gains.}
\label{tab:combined-results}
\begin{center}
\setlength{\tabcolsep}{4pt}
\begin{tabular}{lll | cc | ccc | ccc | cc}
\hline
 & & & \multicolumn{2}{c|}{Reward$\uparrow$} & \multicolumn{3}{c}{Qwen3.5 Win Rate$\uparrow$} & \multicolumn{3}{c|}{GPT-4 Win Rate$\uparrow$} & \multicolumn{2}{c}{RealScore$\uparrow$} \\
\cline{4-5} \cline{6-8} \cline{9-11} \cline{12-13}
Reward & Model & Method & Bas. & Ours & Bas. & \textbf{Ours} & Tie & Bas. & \textbf{Ours} & Tie & Bas. & \textbf{Ours} \\
\hline
\multirow{5}{*}{\raisebox{-0.5ex}{PickScore}}
& SD3.5-M & NFT & 0.869 & 0.866 & 35\% & \cellcolor{tablegray}\textbf{\textcolor{goodcolor}{64\%}} & 1\% & 30\% & \cellcolor{tablegray}\textbf{\textcolor{goodcolor}{67\%}} & 2\% & 0.248 & \cellcolor{tablegray}\textbf{\textcolor{goodcolor}{0.310}} \\
& SD3.5-M & DPO & 0.849 & 0.851 & 28\% & \cellcolor{tablegray}\textbf{\textcolor{goodcolor}{68\%}} & 4\% & 20\% & \cellcolor{tablegray}\textbf{\textcolor{goodcolor}{47\%}} & 33\% & 0.468 & \cellcolor{tablegray}\textbf{\textcolor{goodcolor}{0.472}} \\
& SD3.5-M & AWM & 0.869 & 0.880 & 28\% & \cellcolor{tablegray}\textbf{\textcolor{goodcolor}{72\%}} & 0\% & 23\% & \cellcolor{tablegray}\textbf{\textcolor{goodcolor}{73\%}} & 4\% & 0.283 & \cellcolor{tablegray}{\textcolor{badcolor}{0.269}} \\
& FLUX.2-4B & NFT & 0.905 & 0.910 & 38\% & \cellcolor{tablegray}\textbf{\textcolor{goodcolor}{47\%}} & 15\% & 46\% & \cellcolor{tablegray}\textbf{\textcolor{goodcolor}{51\%}} & 3\% & 0.239 & \cellcolor{tablegray}\textbf{\textcolor{goodcolor}{0.274}} \\
& FLUX.2-4B & AWM & 0.886 & 0.886 & 41\% & \cellcolor{tablegray}\textbf{\textcolor{goodcolor}{59\%}} & 0\% & 38\% & \cellcolor{tablegray}\textbf{\textcolor{goodcolor}{55\%}} & 7\% & 0.277 & \cellcolor{tablegray}\textbf{\textcolor{goodcolor}{0.320}} \\
\hline
\multirow{2}{*}{\raisebox{-0.5ex}{HPSv2}}
& SD3.5-M & NFT & 0.308 & 0.309 & 42\% & \cellcolor{tablegray}\textbf{\textcolor{goodcolor}{58\%}} & 0\% & 47\% & \cellcolor{tablegray}\textbf{\textcolor{goodcolor}{52\%}} & 1\% & 0.218 & \cellcolor{tablegray}\textbf{\textcolor{goodcolor}{0.299}} \\
& FLUX.2-4B & NFT & 0.315 & 0.311 & 36\% & \cellcolor{tablegray}\textbf{\textcolor{goodcolor}{63\%}} & 1\% & 36\% & \cellcolor{tablegray}\textbf{\textcolor{goodcolor}{60\%}} & 4\% & 0.240 & \cellcolor{tablegray}\textbf{\textcolor{goodcolor}{0.274}} \\
\hline
\end{tabular}
\end{center}
\end{table}

Table~\ref{tab:combined-results} reports both MLLM-judged pairwise win rates and forensic realism scores across all configurations. \textbf{1)} On image quality, our regularizer is preferred over the unregularized baseline by both Qwen3.5-35B and GPT-4.1 in all seven settings, spanning two base models, three fine-tuning algorithms, and two reward signals. The two judges agree on preference direction in every case, suggesting that the improvements reflect genuine quality gains rather than evaluator bias. \textbf{2)} On forensic realism, our method improves RealScore in six out of seven configurations. Under AWM, \textbf{\methodname} appears to trade a small amount of detector-based realism for a larger gain in MLLM perceptual quality, suggesting that these axes are not perfectly aligned. \textbf{3)} On reward, our method largely preserves the gains of the unregularized baseline, with PickScore differences of $-0.003 \sim +0.011$ and HPSv2 differences of $-0.004 \sim +0.001$. This suggests that controlling radial inflation does not appear to materially hinder the directional changes that are more reward-aligned in our measurements, consistent with our diagnosis.

Figure~\ref{fig:reward_vs_quality} visualizes the relationship between reward and quality across all configurations. The arrows from baseline to ours are nearly vertical: MLLM quality improves substantially while PickScore remains largely unchanged.

\paragraph{Image-Space Effects of Norm Inflation}
We now test the energy chain predicted in Section~\ref{sec:phenomenon} by measuring the luminance AC energy $E_{\text{lum}}$ and the chroma energy $E_{\text{chroma}}$ of the generated images. Figure~\ref{fig:quality-energy} confirms the prediction: naive RL drives both statistics away from the pretrained reference, inflating $E_{\text{lum}}$ by $24\%$ and $46\%$ in the two settings, and \textbf{\methodname} reduces the magnitude of both deviations. This supports the interpretation that velocity-norm control regulates how much energy is injected into the non-DC image components.

\subsection{Ablations}
\begin{table}[t]
\scriptsize
\renewcommand{\arraystretch}{1.1}
\begin{minipage}[t]{0.48\textwidth}
\centering
\caption{\textbf{Few-step ablation} (FLUX.2-4B, NFT, PickScore). Our regularizer's advantage grows as steps decrease.}
\label{tab:few-step}
\setlength{\tabcolsep}{4pt}
\begin{tabular}{l cc cc cc}
\toprule
 & \multicolumn{2}{c}{RealScore$\uparrow$} & \multicolumn{3}{c}{MLLM Win\%} \\
\cmidrule(lr){2-3} \cmidrule(lr){4-6}
Steps & Bas. & \textbf{Ours} & Bas. & \textbf{Ours} & Tie \\
\midrule
28 (Default) & 0.239 & \cellcolor{tablegray}\textcolor{goodcolor}{\textbf{0.274}} & 38\% & \cellcolor{tablegray}\textcolor{goodcolor}{\textbf{47\%}} & 15\% \\
10 & 0.233 & \cellcolor{tablegray}\textcolor{goodcolor}{\textbf{0.264}} & 39\% & \cellcolor{tablegray}\textcolor{goodcolor}{\textbf{61\%}} & 0\% \\
4  & 0.189 & \cellcolor{tablegray}\textcolor{goodcolor}{\textbf{0.221}} & 40\% & \cellcolor{tablegray}\textcolor{goodcolor}{\textbf{60\%}} & 0\% \\
\bottomrule
\end{tabular}

\end{minipage}
\hfill
\begin{minipage}[t]{0.48\textwidth}
\centering
\caption{\textbf{Early stopping ablation} (FLUX.2-4B, NFT, PickScore). Our method outperforms all earlier checkpoints w/o \textbf{\methodname} .}
\label{tab:early-stop}
\setlength{\tabcolsep}{4pt}
\begin{tabular}{l ccc}
\toprule
Baseline Ckpt. & PickScore$\uparrow$ & RealScore$\uparrow$ & MLLM$\uparrow$ \\
\midrule
Iter 160 & 0.8978 & 0.259 & 6.08 \\
Iter 180 & 0.9010 & 0.252 & 5.99 \\
Iter 200 & 0.9053 & 0.240 & 6.05 \\
\midrule
\rowcolor{tablegray} Ours (200) & \textcolor{goodcolor}{\textbf{0.9103}} & \textcolor{goodcolor}{\textbf{0.274}} & \textcolor{goodcolor}{\textbf{6.16}} \\
\bottomrule
\end{tabular}
\end{minipage}
\end{table}

\paragraph{Few-step robustness.}
Table~\ref{tab:few-step} reports quality metrics as inference steps are reduced from $28$ to $4$. The advantage of our regularizer grows monotonically with fewer steps: the MLLM win-rate gap widens from $9\%$ at $28$ steps to $20\%$ at $4$ steps, while the baseline's RealScore degrades sharply ($0.239$ $\to$ $0.189$) and ours remains comparatively stable. This is consistent with our diagnosis: fewer steps mean larger step sizes in the ODE integrator, amplifying the effect of velocity-norm inflation on discretization error.

\paragraph{Not explained by early stopping.}
A potential confound is that our regularizer simply slows training, and the same quality could be obtained by stopping the baseline earlier. Table~\ref{tab:early-stop} rules this out: we compare baseline checkpoints at iterations $160$, $180$, and $200$ against our method at iteration $200$. Our regularized model achieves higher reward, higher RealScore, and higher MLLM score than \emph{any} earlier baseline checkpoint. The improvement is not accounted for by the early-stopping baselines we test.

\paragraph{Complementarity with KL regularization.}
\begin{figure}[ht]
    \centering
    \begin{minipage}[t]{0.37\textwidth}
        \scriptsize
        \renewcommand{\arraystretch}{1.0}
        \setlength{\tabcolsep}{5pt}
        \begin{center}
            \captionof{table}{\textbf{KL complementarity} (FLUX.2-4B, NFT, PickScore). Our regularizer ($\lambda > 0$) improves RealScore at every KL strength ($\beta_{\text{KL}}$).}
            \label{tab:kl-complementarity}
            \begin{tabular}{cc cc}
                \toprule
                $\beta_{\text{KL}}$ & $\lambda$ & PickScore$\uparrow$ & RealScore$\uparrow$ \\
                \midrule
                0.001 & 0 & 0.9034 & 0.256 \\
                \rowcolor{tablegray} 0.001 & 1.0 & \textcolor{badcolor}{0.8993} & \textcolor{goodcolor}{\textbf{0.329}} \\
                \midrule
                0.0001 & 0 & 0.9004 & 0.270 \\
                \rowcolor{tablegray} 0.0001 & 1.0 & \textcolor{goodcolor}{\textbf{0.9110}} & \textcolor{goodcolor}{\textbf{0.278}} \\
                \midrule
                0 & 0 & 0.8978 & 0.259 \\
                \rowcolor{tablegray} 0 & 1.0 & \textcolor{goodcolor}{\textbf{0.9103}} & \textcolor{goodcolor}{\textbf{0.274}} \\
                \bottomrule
            \end{tabular}
        \end{center}
    \end{minipage}%
    \hfill
    \begin{minipage}[t]{0.30\textwidth}
        \centering
        \vspace{0cm} 
        \includegraphics[width=\textwidth]{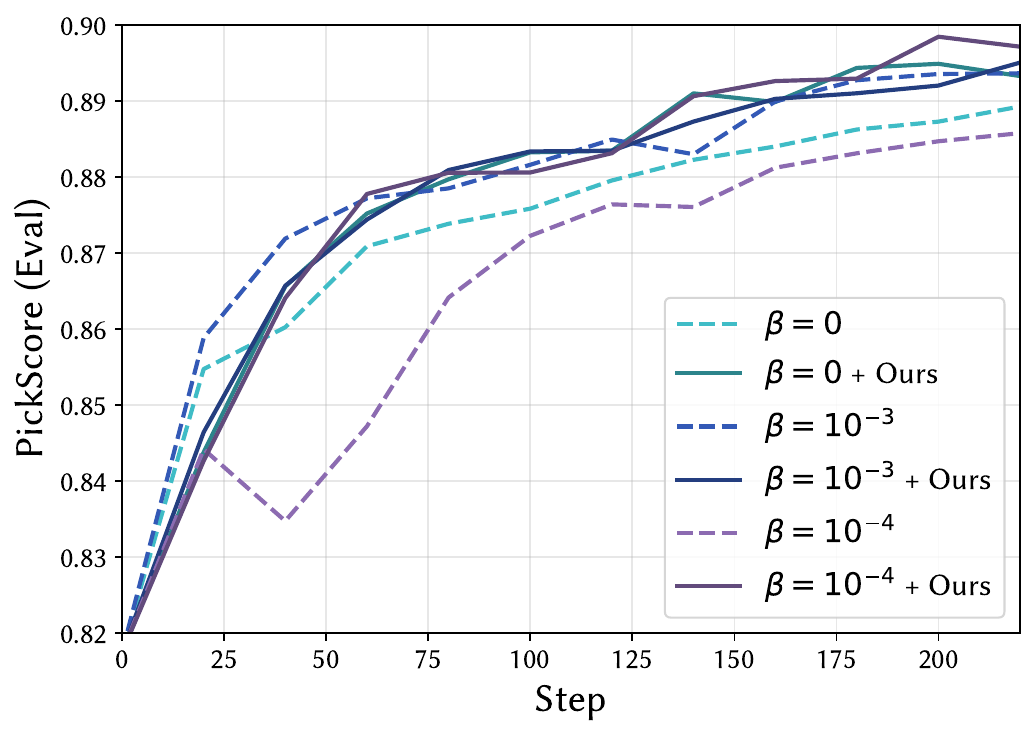}
        \captionof{figure}{Training curve of PickScore evaluation corresponding to Table~\ref{tab:kl-complementarity}.}
        \label{fig:kl-aba-curve}
    \end{minipage}
    \hfill
    \begin{minipage}[t]{0.30\textwidth}
        \centering
        \vspace{0cm} 
        \includegraphics[width=\textwidth]{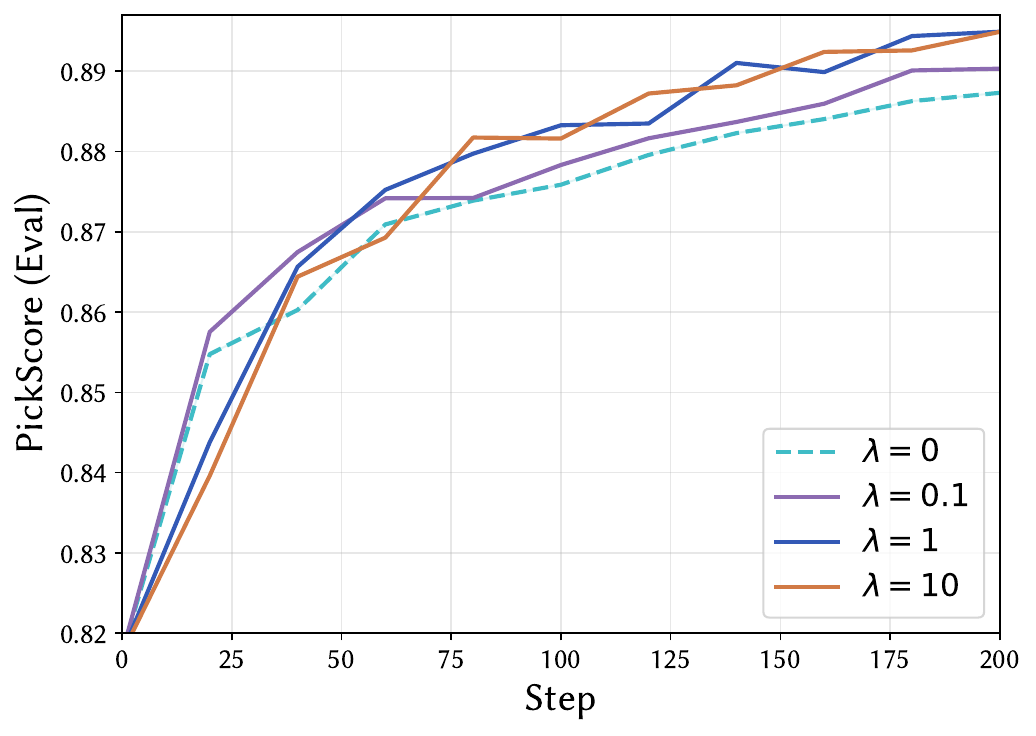}
        \captionof{figure}{PickScore evaluation with different regularizer strengths ($\lambda$).}
        \label{fig:lambda-aba-curve}
    \end{minipage}
\end{figure}

Table~\ref{tab:kl-complementarity} shows that our regularizer is complementary to KL regularization rather than redundant with it. At every level of $\beta_{\text{KL}}$, adding our norm regularization ($\lambda_{\text{VN}} = 1.0$) consistently improves RealScore. Our regularizer targets excess radial growth specifically, whereas KL does not selectively suppress this norm inflation and may also constrain directional changes that appear more reward-aligned in our measurements. The two penalties appear to address distinct failure modes and combine additively in our measurements.

\paragraph{Sensitivity to $\lambda$.}
We vary the regularization strength over $\lambda \in \{0, 0.1, 1, 10\}$ and track PickScore during training. Nonzero regularization consistently outperforms $\lambda=0$, suggesting that controlling radial growth is important. Among the tested values, intermediate strengths work best: $\lambda=1$ achieves the highest score at $200$ iterations, while $\lambda=10$ appears slightly over-regularized early in training. Overall, performance is stable across a broad range of nonzero $\lambda$, indicating that the method is not sensitive to precise tuning.
\section{Conclusion}

RL post-training inflates the per-step velocity norm by $5$ to $15\%$, producing the same artifact family as CFG over-amplification. Unlike CFG, however, RL inflation is trained into the weights: inference-time renormalization fails because the network has co-adapted to the inflated norm, while an adjoint sensitivity analysis shows that suppressing velocity magnitude carries no coherent first-order reward signal at the batch level. Because training-time intervention is both necessary and safe, we propose \methodname, a one-sided hinge penalty on excess velocity norm that composes additively with any velocity-local base loss. Across two base models, three post-training methods, and two reward proxies, \textbf{\methodname} consistently improves perceptual quality and realism while preserving reward, with gains that amplify under few-step inference and are not explained by early stopping or redundant with KL regularization.

\paragraph{Limitations and future work.}
The dynamic origin of the radial inflation remains an open question. Our reward-neutrality claim concerns only first-order signal at the batch level; per-sample radial perturbations do affect reward, but their effects average to zero across prompts. Our analysis applies to velocity-local objectives (NFT, AWM, DPO); extending the norm-budget perspective to trajectory-level objectives such as Flow-GRPO, where gradients flow through likelihood ratios over reverse transitions, is left for future investigation.


\bibliography{iclr2027_conference}
\bibliographystyle{iclr2027_conference}

\newpage
\appendix
\section{Theoretical Analysis of Velocity-Local Post-Training Losses}
\label{sec:appendix-velocity-local}

This appendix provides the formal derivations underlying the velocity-local
property in
Section~\ref{sec:method-velocity-local}. We unify four representative RL post-training
methods (Diffusion-DPO, Diffusion-NFT, AWM, Flow-GRPO) under a single
notation and identify which of them admit a velocity-residual gradient form.


\subsection{Notation}
\label{sec:appendix-notation}

\paragraph{Shared symbols.}
\begin{itemize}[leftmargin=*, itemsep=2pt]
  \item $x_t = (1-t) x_0 + t\,\epsilon$ with $\epsilon \sim \mathcal{N}(0, I)$:
        noisy latent at time $t$.
  \item $v_\theta(x_t, t, c)$: trainable velocity predictor.
  \item $v^{\text{target}} := \epsilon - x_0$: flow-matching target.
  \item $v_{\text{ref}}$: frozen reference velocity field (the pretrained model).
  \item $\omega(t)$: a timestep-dependent weighting.
  \item $J_\theta(x_t, t, c) := \nabla_\theta v_\theta(x_t, t, c)$: per-sample
        velocity Jacobian.
\end{itemize}

The flow-matching pretraining objective is
\begin{equation}
  \mathcal{L}_{\text{FM}}(\theta)
  = \mathbb{E}_{x_0, \epsilon, t, c}\left[
      \omega(t)\, \|v_\theta(x_t, t, c) - v^{\text{target}}\|_2^2
    \right].
\end{equation}

\subsection{Method Decompositions in Velocity Coordinates}
\label{sec:appendix-decompositions}

We now express each method's training signal in the coordinate system above
and verify the velocity-local property of Definition~\ref{def:velocity-local}.

\subsubsection{Diffusion-DPO}

Starting from the Bradley--Terry preference model and reparameterizing the reward
through the diffusion likelihood, the practical Diffusion-DPO loss can be rewritten
in a per-timestep comparison form between a winning sample $x_0^w$ and a losing
sample $x_0^l$:
\begin{equation}
\begin{aligned}
  \mathcal{L}_{\text{DPO}}(\theta)
  = -\mathbb{E}_{t, \epsilon^w, \epsilon^l}
    \log \sigma\Big( -\beta T \omega(\lambda_t) \big[
      &\underbrace{\|v_\theta(x_t^w, t) - v_{\text{ref}}(x_t^w, t)\|_2^2}_{\text{winner deviation}} \\
      -\,& \underbrace{\|v_\theta(x_t^l, t) - v_{\text{ref}}(x_t^l, t)\|_2^2}_{\text{loser deviation}}
    \big] \Big).
\end{aligned}
\end{equation}
The gradient is a sigmoid-weighted sum of two velocity-residual gradients with
target $\tilde{v} = v_{\text{ref}}$, hence velocity-local.

\subsubsection{Diffusion-NFT}

NFT constructs implicit positive and negative policies from the same trainable
parameters:
\begin{align}
  v_\theta^+(x_t, c, t) &= (1-\beta)\, v^{\text{old}}(x_t, c, t) + \beta\, v_\theta(x_t, c, t), \\
  v_\theta^-(x_t, c, t) &= (1+\beta)\, v^{\text{old}}(x_t, c, t) - \beta\, v_\theta(x_t, c, t).
\end{align}
The training objective is a supervised regression on the forward process:
\begin{equation}
  \mathcal{L}_{\text{NFT}}(\theta)
  = \mathbb{E}_{c, x_0, t}\Big[
      r\, \|v_\theta^+(x_t, c, t) - v^{\text{target}}\|_2^2
    + (1-r)\, \|v_\theta^-(x_t, c, t) - v^{\text{target}}\|_2^2
    \Big],
\end{equation}
where $r \in \{0, 1\}$ is a binary optimality label derived from the reward.
The gradient is a reward-weighted FM residual to $v^{\text{target}}$, hence
velocity-local.

\subsubsection{Advantage-Weighted Matching (AWM)}

AWM replaces the per-step reverse-transition likelihood used in DDPO with a
sequence-level policy whose likelihood is approximated through the flow-matching
ELBO. The GRPO-style objective is
\begin{equation}
  \mathcal{J}_{\text{AWM}}(\theta)
  = \mathbb{E}_{c, \{x_0^i\}_{i=1}^G \sim \pi_{\theta_{\text{old}}}}
    \frac{1}{G} \sum_{i=1}^G \left(
      \frac{\hat{\pi}_\theta(x_0^i \mid c)}{\hat{\pi}_{\theta_{\text{old}}}(x_0^i \mid c)} \cdot A_i
      - \beta\, D_{\text{KL}}(\hat{\pi}_\theta \,\|\, \hat{\pi}_{\text{ref}})
    \right).
\end{equation}
Under the FM-ELBO surrogate, the likelihood ratio is estimated through the
difference of two flow-matching losses, and the per-sample gradient simplifies to
\begin{equation}
  \nabla_\theta \mathcal{J}_{\text{AWM}}
  \propto -\nabla_\theta \, \mathbb{E}_t\!\left[
    \omega(t)\, \|v_\theta(x_t, t, c) - v^{\text{target}}\|_2^2
  \right] \cdot A_i.
\end{equation}
The KL term is also instantiated as a velocity-space MSE:
\begin{equation}
  D_{\text{KL}} \propto \omega(t)\, \|v_\theta(x_t, t, c) - v_{\text{ref}}(x_t, t, c)\|_2^2.
\end{equation}

AWM makes the structural symmetry between pretraining and RL explicit: at the
optimization level, the policy update is the same per-timestep FM residual,
modulated only by an advantage weight $A_i$. Both the policy and KL terms are
therefore velocity-local.

\subsubsection{Flow-GRPO}
\label{sec:appendix-flow-grpo}

Flow-GRPO formulates the denoising process as an MDP with state $s_t = (c, t, x_t)$,
action $a_t = x_{t-1}$, and policy equal to the reverse transition
$\pi(a_t \mid s_t) = p_\theta(x_{t-1} \mid x_t, c)$. The objective is a clipped
policy gradient with importance ratio
\begin{equation}
  r_t^i(\theta) = \frac{p_\theta(x_{t-1}^i \mid x_t^i, c)}{p_{\theta_{\text{old}}}(x_{t-1}^i \mid x_t^i, c)}.
\end{equation}

To enable stochastic exploration, Flow-GRPO converts the deterministic ODE into an
SDE and discretizes via Euler--Maruyama, yielding a Gaussian reverse transition
\begin{equation}
  p_\theta(x_{t+\Delta t} \mid x_t, c)
  = \mathcal{N}\!\left( \bar{x}_{t+\Delta t, \theta},\; \sigma_t^2 \Delta t \cdot I \right),
\end{equation}
where the mean $\bar{x}_{t+\Delta t, \theta}$ depends on $v_\theta(x_t, t)$
through the drift term. Because the variance is shared between $\pi_\theta$ and
$\pi_{\theta_{\text{old}}}$, the importance ratio reduces to a function of the
two Gaussian means:
\begin{equation}
  r_t^i(\theta) = \exp\!\left(
    \frac{\|x_{t+\Delta t}^i - \bar{x}_{t+\Delta t, \text{old}}\|^2
        - \|x_{t+\Delta t}^i - \bar{x}_{t+\Delta t, \theta}\|^2}
        {2\sigma_t^2 \Delta t}
  \right).
\end{equation}
The gradient flows through this trajectory-level likelihood ratio rather than
through a per-timestep velocity residual; $v_\theta$ enters only indirectly
through the Gaussian transition mean. Flow-GRPO therefore violates
Definition~\ref{def:velocity-local}, and we treat Flow-GRPO as out of scope for this work.

\subsection{Summary}

\begin{table}[h]
\centering
\caption{Classification of post-training methods by velocity-locality
(Definition~\ref{def:velocity-local}).}
\label{tab:velocity-local-classification}
\small
\begin{tabular}{l c p{0.55\linewidth}}
\toprule
Method & Velocity-local? & Reason \\
\midrule
NFT       & \cmark & Gradient is a reward-weighted sum of FM residuals to $v^{\text{target}}$. \\
AWM       & \cmark & FM-ELBO surrogate reduces the policy update to an advantage-weighted FM residual; KL is also a velocity MSE to $v_{\text{ref}}$. \\
DPO       & \cmark & Sigmoid-weighted local quadratic deviations $\|v_\theta - v_{\text{ref}}\|_2^2$. \\
Flow-GRPO & \xmark & Gradient flows through a trajectory-level policy ratio; $v_\theta$ enters only via the Gaussian transition mean. \\
\bottomrule
\end{tabular}
\end{table}

\newpage
\section{Image-Space Energy Statistics} \label{sec:appendix-energy-stats}

We detail the two statistics used in Section~\ref{sec:phenomenon} and Figure~\ref{fig:quality-energy}. Decoded images are normalized to $[0,1]$ per RGB channel and converted to $\mathrm{YC_bC_r}$ following the BT.709 convention,
\begin{equation}
  Y = 0.2126 R + 0.7152 G + 0.0722 B, \qquad
  C_b = \frac{B - Y}{1.8556}, \qquad
  C_r = \frac{R - Y}{1.5748},
\end{equation}
so that $Y$ carries luminance and $(C_b, C_r)$ carry chrominance. For an image of resolution $H \times W$ we compute the per-image spatial second central moments
\begin{equation}
  E_{\text{lum}} := \operatorname{Var}[Y] = \frac{1}{HW}\sum_{x, y}\big(Y(x, y) - \bar{Y}\big)^2,
  \qquad
  E_{\text{chroma}} := \operatorname{Var}[C_b] + \operatorname{Var}[C_r],
\end{equation}
and average them over the evaluation set. Since subtracting the channel mean removes exactly the DC coefficient of the 2D discrete Fourier transform, Parseval's identity gives
\begin{equation}
  E_{\text{lum}} = \frac{1}{(HW)^2}\sum_{(k_x, k_y) \neq (0,0)} \big|\hat{Y}(k_x, k_y)\big|^2 ,
\end{equation}
i.e., $E_{\text{lum}}$ is the total luminance energy summed over all non-DC spatial frequencies, and $E_{\text{chroma}}$ is the same quantity for the two chroma channels. Both statistics are thus invariant to a global brightness or color offset and respond only to how much structure the image places around its own mean.

\begin{figure}[h]
\centering
\includegraphics[width=0.8\linewidth]{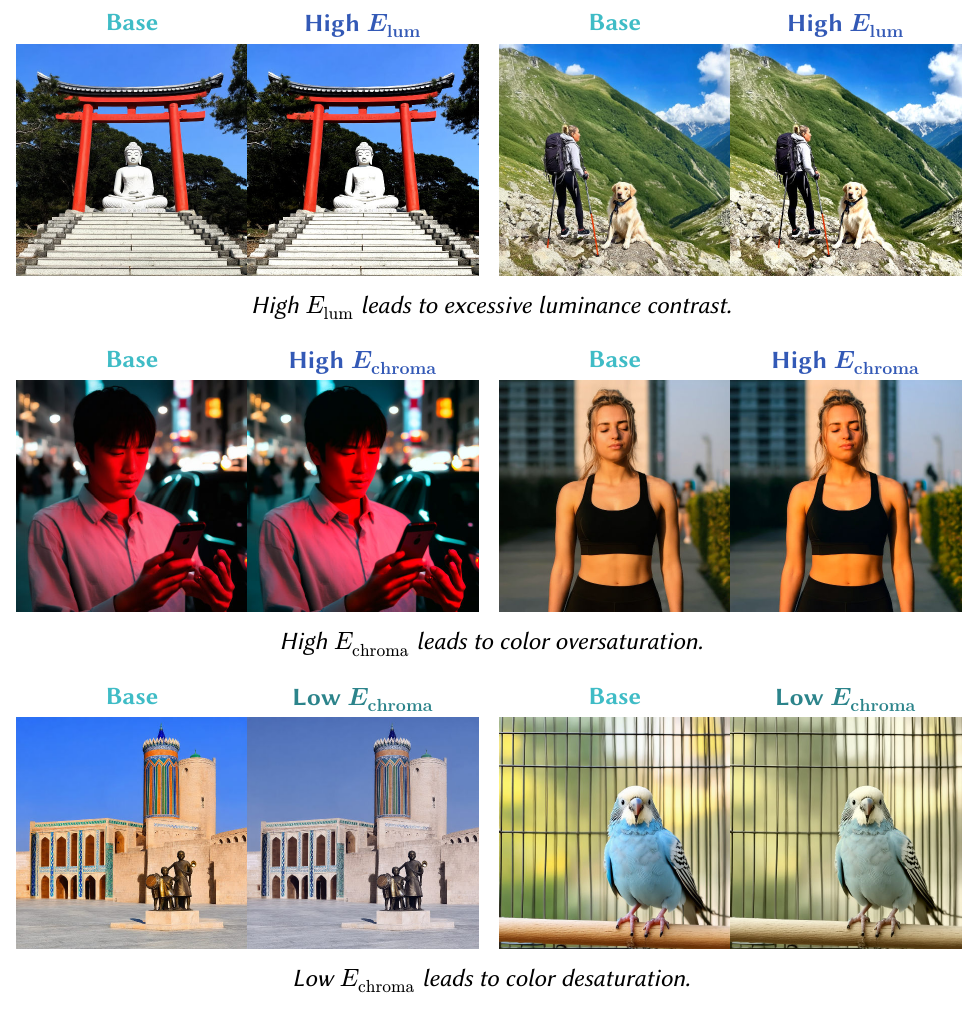}
\vspace{-10pt}
\caption{Examples of high/low $E_\text{lum}$ and $E_\text{chroma}$ images.}
\label{fig:img_stat_examples}
\end{figure}

\newpage

\section{Training Details} \label{sec:appendix-training-details}

Table~\ref{tab:app-hparams-main} provides the full training hyperparameters for each (model, method, reward) combination. Table~\ref{tab:app-hparams-method} details method-specific hyperparameters, and Table~\ref{tab:app-hparams-opt} lists shared optimization settings. For detailed implementation, see Flow-Factory~\cite{Flow-Factory} codes.

\begin{table}[h]
\scriptsize
\centering
\caption{\textbf{Main experiments.} Full hyperparameters for each (model, method, reward) combination. Here \textit{Train Steps} means the number of sampling steps during rollout, and \textit{GS} means group size.}
\label{tab:app-hparams-main}
\begin{tabular}{l l l | c c c | c c c c}
\toprule
Model & Method & \textbf{\methodname} & LR & LoRA $r$ & $\lambda$ & Train Res. & Train Steps & GS & Batch/dev   \\
\midrule
\multirow{6}{*}{SD3.5-M}
& NFT$_{\text{PickScore}}$ & \xmark & $1\!\times\!10^{-4}$ & 32 & -- & 512 & 14 & 16 & 8 \\
& NFT$_{\text{PickScore}}$ & \cmark & $1\!\times\!10^{-4}$ & 32 & 1.0 & 512 & 14 & 16 & 8  \\
& NFT$_{\text{HPSv2}}$ & \xmark & $1\!\times\!10^{-4}$ & 32 & -- & 512 & 14 & 16 & 8  \\
& NFT$_{\text{HPSv2}}$ & \cmark & $1\!\times\!10^{-4}$ & 32 & 1.0 & 512 & 14 & 16 & 8  \\
& AWM$_{\text{PickScore}}$ & \xmark & $3\!\times\!10^{-4}$ & 32 & -- & 512 & 14 & 16 & 8  \\
& AWM$_{\text{PickScore}}$ & \cmark & $3\!\times\!10^{-4}$ & 32 & 0.01 & 512 & 14 & 16 & 8  \\
& DPO$_{\text{PickScore}}$ & \xmark & $1\!\times\!10^{-5}$ & 32 & -- & 512 & 40 & 4 & 24  \\
& DPO$_{\text{PickScore}}$ & \cmark & $1\!\times\!10^{-5}$ & 32 & 1.0 & 512 & 40 & 4 & 24  \\
\midrule
\multirow{4}{*}{FLUX.2-4B}
& NFT$_{\text{PickScore}}$ & \xmark & $1\!\times\!10^{-4}$ & 64 & -- & 384 & 8 & 16 & 2  \\
& NFT$_{\text{PickScore}}$ & \cmark & $1\!\times\!10^{-4}$ & 64 & 1.0 & 384 & 8 & 16 & 2  \\
& NFT$_{\text{HPSv2}}$ & \xmark & $1\!\times\!10^{-4}$ & 64 & -- & 384 & 8 & 16 & 2 \\
& NFT$_{\text{HPSv2}}$ & \cmark & $1\!\times\!10^{-4}$ & 64 & 1.0 & 384 & 8 & 16 & 2  \\
& AWM$_{\text{PickScore}}$ & \xmark & $3\!\times\!10^{-4}$ & 64 & -- & 384 & 8 & 16 & 1  \\
& AWM$_{\text{PickScore}}$ & \cmark & $3\!\times\!10^{-4}$ & 64 & 0.01 & 384 & 8 & 16 & 1 \\
\bottomrule
\end{tabular}
\end{table}

\begin{table}[h]
\scriptsize
\centering
\caption{\textbf{Method-specific hyperparameters.}}
\label{tab:app-hparams-method}
\begin{tabular}{l l l l l l l}
\toprule
Method & Adv.~Agg. & Timestep Sampler & $\#t_{\text{train}}$ & Time Shift & Adv.~Clip & Other \\
\midrule
NFT & sum / gdpo & discrete & 4 (SD3.5) / 2 (FLUX) & 3.0 & 5.0 & EMA decay 0.5 \\
AWM & sum & discrete\_wo\_init & 6 (SD3.5) / 4 (FLUX) & 3.0 & 1.0 & ghuber power 0.25; clip 1.0 \\
Online-DPO & -- & logit\_normal & 1 & 3.0 & -- & $\beta=100$; pair best-vs-worst \\
\bottomrule
\end{tabular}
\end{table}

\begin{table}[h]
\small
\centering
\caption{\textbf{Shared optimization settings.}}
\label{tab:app-hparams-opt}
\begin{tabular}{l c c c c c}
\toprule
Setting & Optimizer & Weight Decay & Betas & $\epsilon$ & Max Grad Norm \\
\midrule
All runs & AdamW & $1\!\times\!10^{-4}$ & $(0.9, 0.999)$ & $1\!\times\!10^{-8}$ & 1.0 \\
\bottomrule
\end{tabular}
\end{table}
\newpage
\section{MLLM Evaluation Prompts} \label{sec:appendix-mllm-prompts}

\begin{promptbox}[title={Image Realism Judge Prompt}]
You are an expert "Image Realism Judge".

Input:
- image_A: the first image
- image_B: the second image
{prompt_section}

Objective:
Compare the two images and determine which one appears more realistic (i.e., more like a genuine, unmanipulated photograph of a real-world scene).

Evaluation Criteria (apply in order of importance):
1. Physical plausibility: lighting, shadows, reflections, perspective, and scale must be consistent with real-world physics.
2. Texture & material fidelity: surfaces (skin, fabric, metal, wood, etc.) should exhibit natural micro-detail, noise, and variation. Penalize images that look overly smooth, plastic, painted, or unnaturally "oily/greasy"
3. Edge & boundary coherence: object boundaries should be natural; look for blurring halos, jagged masks, unnatural cutouts, or excessive / artificial sharpening (e.g., crisp white/black halos around edges, unnaturally hard transitions)
4. Color & tone consistency: global and local color grading should be coherent; watch for inconsistent saturation, clipped highlights, or artificial color casts.
5. Semantic coherence: all depicted objects, scenes, and interactions must be logically plausible (e.g., no floating objects, impossible reflections, anatomical errors).
6. Artifact detection: check for common AI-generated artifacts: repeated patterns, watermarks, noise inconsistencies, duplicate elements, ghost limbs, distorted text.

Judgment Protocol:
A) For each image, list specific realism strengths and weaknesses with concrete visual evidence.
B) Score each image on a 1–10 realism scale (10 = indistinguishable from a real photograph).
C) Decide which image is more realistic: "A" or "B". If both are equally realistic, output "tie".
D) Provide a brief justification focusing on the most decisive differences.

Output format:
Return STRICT JSON only (no markdown, no extra text), using this schema:

{
  "image_A_analysis": {
    "realism_score": <int 1-10>,
    "strengths": ["..."],
    "weaknesses": ["..."],
    "key_evidence": ["..."]
  },
  "image_B_analysis": {
    "realism_score": <int 1-10>,
    "strengths": ["..."],
    "weaknesses": ["..."],
    "key_evidence": ["..."]
  },
  "winner": "A" | "B" | "tie",
  "justification": "..."
}

Now compare these two images and judge which is more realistic.
\end{promptbox}

\newpage
\section{Prompts for Main-Paper Figures}
\label{sec:appendix-prompts}

This appendix lists the text prompts used to generate qualitative examples that appear in the main paper. We group prompts by figure for ease of reference. All the prompts are sampled from HPDv3~\cite{HPSv2}.
\subsection{Teaser}
\label{sec:appendix-prompts-teaser}

\begin{promptbox}[title={
Tropical coastline scene.
}]
Captured from an elevated perspective, the image showcases a tropical paradise, where lush greenery meets the vibrant turquoise ocean. The coastline is marked by a cliff face covered in dense foliage, interspersed with large, rocky formations jutting out into the sea.  A pristine white sand beach curves along the coast, gently kissed by the foamy waves. Palm trees sway rhythmically along the shoreline, casting shadows over what appears to be a small building with a thatched roof. The ocean's color gradient is striking, transitioning from a deep azure in the distance to a clear turquoise near the shore, revealing sandy patches beneath the surface.  The scene evokes a sense of tranquility and seclusion, ideal for a tropical getaway. The composition highlights the natural beauty of the landscape, emphasizing the contrast between the verdant vegetation, the golden sand, and the crystal-clear waters.
\end{promptbox}
\begin{promptbox}[title={
Bird over mountain vista.
}]
The image captures a breathtaking mountain landscape under a bright, clear sky. A bird, silhouetted against the light, is the focal point of the composition. It appears to be perched on a weathered wooden railing, wings partially extended as if ready to take flight. The bird's dark feathers contrast sharply with the radiant sunlight behind it, highlighting its form.  In the background, a range of majestic mountains stretches towards the horizon. The peaks are rendered in soft shades of blue and gray, fading into the distance as they recede. The terrain below the mountains is rocky and sparsely vegetated, suggesting a rugged, high-altitude environment.  The wooden railing in the foreground adds a sense of depth and perspective to the image. The warm tones of the wood provide a textural contrast to the cool blues and grays of the mountains. The overall effect is a serene and awe-inspiring scene that captures the beauty and grandeur of nature.
\end{promptbox}
\begin{promptbox}[title={
Tree silhouette at sunset.
}]
The prompt in English translates to:  This image is a detailed computer-generated artwork depicting a tranquil sunset scene. The foreground shows a nearly black horizon that gradually transitions into blue-gray tones as the distance increases. At the center of the picture stands a slender tree, with its sparse leaves forming a striking silhouette against the vibrant sky. The tree's trunk is thin and long, with sparsely distributed leaves on its branches, creating a delicate contrast with the colorful background.
\end{promptbox}

\subsection{Main Results}
\label{sec:appendix-prompts-main-results}
\textbf{SD3.5 + PickScore + NFT.}
\begin{promptbox}[title={
Person with smartphone, red-lit urban night.
}]
The image captures a person holding a smartphone in an urban setting, likely at night. The individual is the primary focus, positioned slightly off-center, wearing a light-colored button-down shirt. The red lighting casts a distinct glow on their face and parts of their clothing, creating a dramatic contrast.  Their short, dark hair is casually styled, framing their face. The person's expression is somewhat neutral, perhaps thoughtful or observant. The smartphone is held in their hand, suggesting they may be texting, browsing, or waiting for something.  The background is blurred, indicating a shallow depth of field. It consists of city streets with blurry pedestrians, vehicles, and building lights. A black car, possibly a taxi, is faintly visible behind the person, adding to the urban feel. The bokeh from the lights creates a soft, dreamy effect. The overall tone is modern and slightly mysterious due to the lighting and blurred background.
\end{promptbox}
\begin{promptbox}[title={
3D sculpted book landscape.
}]
The image presents an open book, standing upright with its pages forming a vibrant, three-dimensional landscape. The pages have been sculpted to resemble a steep, rocky terrain, teeming with lush greenery. Verdant foliage, including moss, ferns, and small trees, clings to the rocky surfaces, creating a miniature, intricate ecosystem.   A lone figure, appearing small in the grand scale of the scene, is walking through the landscape. This individual is set against a backdrop of soft light, which suggest the break of dawn, or a hidden clearing ahead. The lighting adds to the dreamy, surreal quality of the image, enhancing the feeling of entering a fantastical world.   The book's cover and visible pages have a rustic, weathered look, adding to the sense of history and discovery. The scene sits on what seems to be a sandy or earthy surface, grounding the book and its internal world in a tangible reality. The overall composition invokes a sense of wonder, inviting the viewer to explore the boundless possibilities contained within the pages of a book.
\end{promptbox}
\begin{promptbox}[title={
Pumpkin patch in autumn.
}]
The image shows a field of pumpkins, closely packed together, suggesting a pumpkin patch or harvest scene. The pumpkins vary in size and color, with most appearing in shades of orange. Their textures seem rough, and the shapes are somewhat irregular. The stems are thick and curved, appearing dark green or brown. Some green grass blades are visible between the pumpkins in the foreground.  The focus is on the pumpkins in the front, making them appear larger, while the pumpkins in the background are slightly out of focus, creating depth and perspective in the image. The lighting seems soft and natural, enhancing the warm, earthy tones of the pumpkins. The overall impression is one of abundance and autumn.
\end{promptbox}

\textbf{SD3.5 + PickScore + AWM.}
\begin{promptbox}[title={
Buddha statue with torii gate, twilight.
}]
The image depicts a tranquil outdoor scene featuring a prominent white Buddha statue seated with legs crossed and hands resting on its knees. The statue, made of marble or similar material, is situated on a small hill surrounded by several trees. In the foreground, there are three concrete steps leading up to a red torii gate, which is a traditional Japanese wooden entrance arch commonly found at Shinto shrines. The torii gate consists of two vertical posts and two horizontal beams, with the top beam being longer than the bottom one. The sky transitions from deep blue near the top to a lighter shade of blue towards the horizon, suggesting that the time of day might be either dusk or dawn. Overall, the composition evokes a serene and culturally rich atmosphere.
\end{promptbox}
\begin{promptbox}[title={
Woman on steps, urban mural background.
}]
The image features a woman sitting on a concrete surface, likely a step or platform, with stone steps rising behind her. She is wearing a red and black plaid flannel shirt and dark jeans, paired with white Converse sneakers. Her dark hair is styled short, and she is wearing brown sunglasses.   To her left is a brown leather backpack. The background includes a building with a faded mural or painting. A tree partially obscures the building, adding a touch of nature to the urban setting. The overall impression is one of a casual moment, perhaps a pause during exploration or travel. The lighting is bright, suggesting a sunny day, and the composition is balanced, with the woman as the central focus and the surrounding elements adding context and interest.
\end{promptbox}
\begin{promptbox}[title={
Graffiti wall, woman and alien character.
}]
This painting depicts a vibrant and detailed graffiti wall that blends elements of cartoon and realism. At the center of the image is a woman with long wavy blonde hair, wearing a black tank top and brown shorts, her arms crossed as she stares directly at the viewer. To her left is a comical green alien figure with a cylindrical head, a large single eye, and a spray paint can in hand, seemingly in the process of creating more graffiti. On the right side of the wall is a complex black-and-white design featuring geometric shapes and mask-like faces, adding to the intricacy of the work. The background is a light blue, dotted with additional graffiti elements. In the foreground, a man dressed in black sits on the sidewalk, engrossed in his phone, seemingly oblivious to the bustling scene behind him. Overall, the composition is rich in color and detail, capturing the essence of urban street art.
\end{promptbox}

\textbf{SD3.5 + PickScore + AWM + 10 Steps.}

\begin{promptbox}[title={
Abstract painted landscape, stylized village.
}]
Certainly! Here is a description of the image:  The image presents an abstract landscape painting in a vibrant color palette. In the foreground, a stylized field or body of water dominates, rendered in shades of purple and blue with visible brushstrokes.   Centered in the midground is a cluster of simplified, white-walled buildings or structures, reminiscent of a small village or town. These buildings are clustered together, creating a compact silhouette against the sky. Above the structures are stylized trees with pointy shapes that create a decorative feel.  The background is a blend of warm, muted tones, suggesting a sunset or sunrise. The overall composition is balanced, with the dark field in the foreground grounding the lighter, more colorful elements above. The style is simplified and somewhat naive, emphasizing color and shape over realistic representation.
\end{promptbox}
\begin{promptbox}[title={
Solitary figure on rocky mountain precipice.
}]
The image depicts a solitary figure standing on the edge of a rocky precipice, with a towering mountain range as the backdrop. The person, dressed in dark clothing, is positioned on a flat, rocky outcrop, looking out towards the majestic landscape.  The mountains are rugged and imposing, with sharp, jagged peaks that reach towards the sky. The upper portion of the highest peak is shrouded in mist or clouds, adding an element of mystery and grandeur to the scene. The slopes of the mountain are steep and rocky, with visible scree and paths, indicating the challenging terrain.  The foreground features a rocky landscape with patches of vegetation, providing a textural contrast to the smooth, pale slopes of the mountain. The overall tone of the image is one of awe and solitude, emphasizing the smallness of the individual in the face of nature's immense power and beauty. The muted color palette and the presence of mist suggest a cool, perhaps even slightly cold, climate.
\end{promptbox}
\begin{promptbox}[title={
Hanukkah gifts, snow globe with dreidel.
}]
This photo captures a heartwarming scene of Hanukkah gifts on a table. On the left, a charming snow globe featuring a dreidel is the focal point. The globe is clear, encapsulating a blue and white dreidel with Hebrew letters on each side, set against a blurred backdrop. The base of the globe is dark blue, adorned with white Jewish stars and swirls, trimmed with a delicate white border.  To the right, a gift is wrapped in shiny blue paper, decorated with a festive pattern of white Hanukkah-themed designs, including stars and menorahs. The wrapping adds a touch of elegance and anticipation to the scene.  The table itself is textured, adding depth to the composition. The background is softly blurred, suggesting an outdoor setting, possibly a patio or garden. Overall, the image conveys the warmth and joy of the Hanukkah season, emphasizing the beauty of tradition and the excitement of gift-giving. The lighting is soft and natural, creating a cozy and inviting atmosphere.
\end{promptbox}

\textbf{FLUX.2 + PickScore + NFT.}
\begin{promptbox}[title={
Couple on skateboard, sepia road at dusk.
}]
The image captures a warm, intimate moment between a young couple sitting on a skateboard in the middle of a deserted road. The color palette leans towards sepia tones, giving the scene a nostalgic, almost vintage feel.  The man wears a dark long-sleeved shirt with a patterned sleeve and a backwards baseball cap. He's seated with his back mostly exposed, one arm wrapped around the woman, as they lean their heads towards each other. The woman is dressed in a denim jacket and dark pants. Her hair is dark and flows down her back as she leans slightly into the man.  They sit on a skateboard, casually placed on the asphalt road. The road stretches into the distance, flanked by golden, grassy hills. The sky is a soft, muted yellow, giving the impression of either dawn or dusk. The overall feeling is one of peaceful companionship and shared solitude.
\end{promptbox}
\begin{promptbox}[title={
Metallic sculpture, Milky Way backdrop.
}]
The image showcases a mesmerizing juxtaposition of earthly art and celestial wonder. In the foreground, a metallic, human-like sculpture stands on a barren, cracked landscape. The figure appears to be introspective, its head tilted downwards, seemingly lost in contemplation. The reflective surface of the sculpture gleams subtly, picking up the faint light from the night sky.  Dominating the background is a breathtaking view of the Milky Way, stretching like a luminous river across the deep blue canvas of the night. The core of our galaxy blazes with countless stars, its swirling patterns creating a sense of depth and immensity. A faint reddish hue hints at distant nebulae. The stars shimmer and scatter across the sky, creating a magical, almost surreal atmosphere. The stark contrast between the grounded sculpture and the boundless expanse of the cosmos invites reflection on humanity's place in the universe.
\end{promptbox}
\begin{promptbox}[title={
Man in modern chair, whiskey glass, airy interior.
}]
The image shows a well-dressed man sitting in a stylish, modern chair, exuding an air of relaxed sophistication. He wears a light grey, button-down shirt, neatly tucked into navy-blue trousers, complemented by black leather shoes. His dark hair is neatly styled. The man is holding a glass of amber-colored liquid, possibly whiskey, in his hands, but not drinking, his eyes are gazed upwards.  The chair he is seated in has a sleek, black leather cushion and a minimalist metal frame. His legs are crossed, adding to the casual and comfortable posture. Beside him, a wooden table holds a smartphone, suggesting a moment of disconnecting from the digital world.  The image has a bright, airy feel. The background is somewhat blurred, with hints of modern architectural elements, further emphasizing the contemporary and sophisticated atmosphere. The soft lighting enhances the scene, creating a sense of calm and introspection.
\end{promptbox}

\textbf{FLUX.2 + HPSv2 + NFT.}
\begin{promptbox}[title={
Camper at dusk, lantern and portable stove.
}]
The image captures a serene moment of a camper enjoying the peaceful ambiance of an outdoor setting as dusk settles in. A young man is seated comfortably in a portable camping chair, sporting a casual grey hoodie and dark hiking pants. His gaze is directed downwards, giving off a contemplative vibe.  Next to him, a small, portable stove is lit, casting a warm glow. Nearby, a makeshift table setup consisting of a black storage box topped with a wooden surface displays various camping essentials. A lantern provides ample light, illuminating the immediate area.  In the background, a tent stands as a simple shelter. The surrounding landscape features lush greenery and palm trees under a twilight sky. The composition is balanced and creates a sense of tranquility and the simple pleasures of camping.
\end{promptbox}
\begin{promptbox}[title={
Couple on wet beach, ocean backdrop, golden hour.
}]
The image captures a serene scene of a couple lying on a sandy beach with a backdrop of the ocean and a partly cloudy sky. The couple is positioned in the foreground, their bodies aligned horizontally. The man is wearing patterned swim trunks, while the woman is in a pink bikini. They appear relaxed and content, with their heads close together, creating an intimate atmosphere.  The beach appears to be wet, reflecting the soft, diffused light, which adds to the tranquility of the scene. In the background, the ocean stretches out to the horizon, where distant landmasses can be seen. The sky transitions from a deep blue at the top of the frame to a lighter hue near the horizon, with a few fluffy clouds adding depth and texture to the composition. The overall tone of the image is peaceful and romantic.
\end{promptbox}
\begin{promptbox}[title={
Solitary figure on rocky precipice, misty peaks.
}]
The image depicts a solitary figure standing on the edge of a rocky precipice, with a towering mountain range as the backdrop. The person, dressed in dark clothing, is positioned on a flat, rocky outcrop, looking out towards the majestic landscape.  The mountains are rugged and imposing, with sharp, jagged peaks that reach towards the sky. The upper portion of the highest peak is shrouded in mist or clouds, adding an element of mystery and grandeur to the scene. The slopes of the mountain are steep and rocky, with visible scree and paths, indicating the challenging terrain.  The foreground features a rocky landscape with patches of vegetation, providing a textural contrast to the smooth, pale slopes of the mountain. The overall tone of the image is one of awe and solitude, emphasizing the smallness of the individual in the face of nature's immense power and beauty. The muted color palette and the presence of mist suggest a cool, perhaps even slightly cold, climate.
\end{promptbox}

\end{document}